\documentclass[table]{reportstyle/report}

\usepackage[utf8]{inputenc} %
\usepackage[T1]{fontenc}    %

\usepackage{amssymb}
\usepackage{fdsymbol}   %

\usepackage{url}            %
\usepackage{booktabs}       %
\usepackage{amsfonts}       %
\usepackage{nicefrac}       %
\usepackage{microtype}      %
\usepackage{amsmath}
\usepackage{dialogue}
\usepackage{pifont}%

\usepackage{siunitx}
\usepackage{graphicx}

\usepackage{wrapfig}
\usepackage{enumitem}
\setlist[itemize]{left=3pt, before=\vspace{-2pt}} %
\usepackage{multirow}
\usepackage{adjustbox}
\usepackage{pifont}
\usepackage{caption}
\usepackage{hyperref}

\usepackage{nicematrix}
\usepackage{listings}

\usepackage{hyperref}
\usepackage{url}
\usepackage{graphicx}

\usepackage{booktabs}
\usepackage{multirow}
\usepackage[table]{xcolor}
\usepackage{graphicx}
\usepackage{hhline}
\usepackage{caption}
\usepackage{adjustbox}

\usepackage{tabularx}
\usepackage{array}

\usepackage{booktabs}
\usepackage{makecell}
\usepackage{subcaption}

\newcolumntype{C}{>{\centering\arraybackslash}X}

\makeatletter
\renewcommand{\sectionautorefname}{\S\@gobble}
\renewcommand{\subsectionautorefname}{\S\@gobble}
\renewcommand{\subsubsectionautorefname}{\S\@gobble}
\renewcommand{\appendixautorefname}{\S\@gobble}
\makeatother

\title{When to Retrieve, When to Stay: Uncertainty-Aware Temporal Evidence Allocation for Streaming Video-LLMs}

\newcommand\extrafootertext[1]{%
    \bgroup
    \renewcommand\thefootnote{\fnsymbol{footnote}}%
    \renewcommand\thempfootnote{\fnsymbol{mpfootnote}}%
    \footnotetext[0]{#1}%
    \egroup
}

\authorOne[1\dagger]{Xiang Hu}
\authorOne[1\dagger]{Jiazuo Yu}
\authorOne[1]{Lu Zhang}
\authorOne[1]{Yunzhi Zhuge}
\authorOne[1]{Huchuan Lu}

\affiliation[1]{IIAU Lab, Dalian University of Technology}

\contribution[\dagger]{%
  Equal Contributions\protect\\[8pt]
  Contact Email: \email{zgyz@dlut.edu.cn}%
}

\abstract{

Streaming video understanding requires  Video Large Language Models (Video-LLMs) to reason over continuous visual streams under causal constraints. 
As the visual history grows, a bounded visual-processing budget requires evidence selection that balances temporal recency with query relevance. 
Recent-only selection excludes potentially relevant historical evidence, whereas Semantic-only retrieval can displace useful recent context when relevance scores are ambiguous. 
We introduce WRWS (When to Retrieve, When to Stay), a training-free framework for uncertainty-adaptive evidence allocation. 
A lightweight external vision-language encoder scores query relevance across the observed history, while an adaptive allocation module uses the normalized entropy of the similarity distribution as a proxy for retrieval uncertainty. 
WRWS favors semantic retrieval when relevance cues are reliable and strengthens the recency prior under uncertainty. 
Following a retrieve-first, encode-later pipeline, WRWS selects evidence before target-model visual encoding, such that only the selected observations are processed by the costly target Video-LLM. 
Experiments across four Video-LLM families and multiple model scales demonstrate competitive accuracy on StreamingBench and OVO-Bench. 
In our efficiency evaluation, WRWS reduces average vision-to-answer time to 47.93\% of the state-of-the-art method.
Code will be released.

}

\begin{document}

\maketitle

\section{Introduction}

Multimodal large language models (MLLMs) have advanced video understanding~\citep{llavaonevision,qwen25vl}, yet deploying them on continuously arriving video streams remains challenging.
Unlike offline video understanding, streaming inference must answer queries using only observations available up to the query time, while potentially drawing on an increasingly long history~\citep{lin2024streaming,niu2025ovo}.
Processing this history with the target Video-LLM is expensive, even though only a subset of observations may ultimately be useful for a given query.
This raises a central question: \emph{under a bounded budget for expensive target-model visual processing, which causally available observations should receive target-model visual encoding?}

\begin{figure}[t]
    \centering
    \includegraphics[width=.99\linewidth]{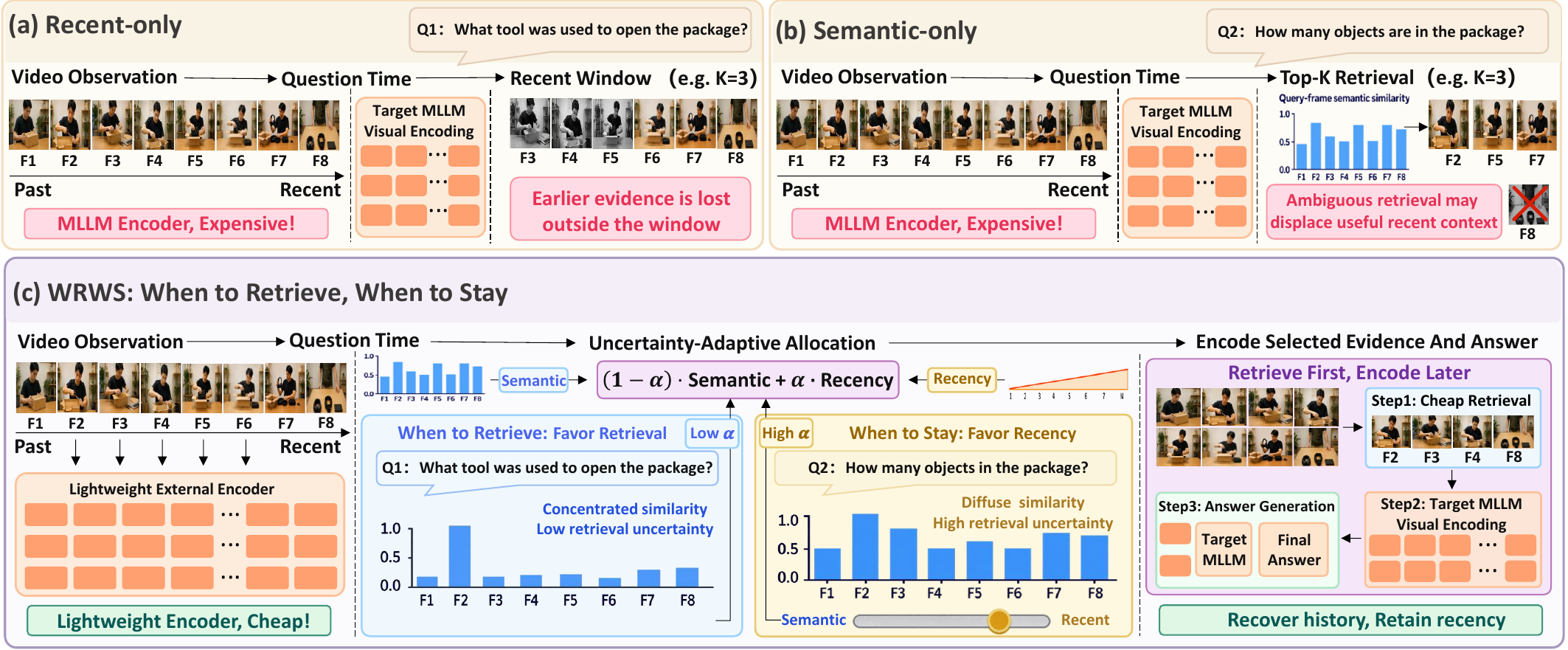}
    \vspace{-2mm}
    \caption{
    Motivation and key idea of WRWS.
    Under a bounded target-model visual-processing budget,
    \textbf{(a)} Recent-only selection may lose useful earlier evidence, while
    \textbf{(b)} Semantic-only retrieval may displace useful recent context when similarity is ambiguous.
    \textbf{(c)} WRWS adapts the semantic--recency trade-off to retrieval uncertainty and follows \emph{retrieve first, encode later}, selecting evidence with lightweight representations before expensive target-model visual encoding.
    }
    \vspace{-6mm}
    \label{fig:teaser}
\end{figure}

A natural training-free strategy is to retain the most recent observations~\citep{simplestream}.
As illustrated in Fig.~\ref{fig:teaser}(a), such Recent-only selection provides a strong inductive bias for queries about the current scene, but earlier evidence becomes inaccessible once it falls outside the retained window.
Expanding the window preserves more history, but also increases the visual-token load and the associated target-model processing cost.
More fundamentally, a fixed Recent-only policy allocates the available budget solely according to temporal proximity, irrespective of which part of the observed history is relevant to the current query.
Semantic retrieval offers a complementary strategy by selecting query-relevant observations from the causally available history~\citep{rekv,ning2025livevlm}.
However, as illustrated in Fig.~\ref{fig:teaser}(b), when query--observation similarities are ambiguous or insufficiently discriminative, semantic ranking provides a weak selection signal, and relying solely on similarity may displace otherwise useful recent context.
The challenge is therefore not to choose recency or retrieval globally, but to determine \emph{how strongly semantic relevance should override the recency prior for each query under a fixed evidence budget}.

These challenges are supported by two empirical observations in Sec.~\ref{sec:empirical_observations}.
First, target-model visual processing constitutes a major component of the measured vision-to-answer time (VTAT) runtime, motivating evidence selection before expensive target-model encoding.
Second, Recent-only and Semantic-only exhibit substantial query-level complementarity under the same evidence budget, suggesting that no fixed policy is uniformly preferable across queries.
%

We therefore introduce \textbf{WRWS (When to Retrieve, When to Stay)}, a training-free framework for \emph{adaptive temporal evidence allocation under a bounded target-model visual-processing budget}.
As illustrated in Fig.~\ref{fig:teaser}(c), WRWS couples uncertainty-adaptive evidence allocation with a \emph{retrieve-first, encode-later} pipeline.
As observations arrive, a lightweight external vision encoder produces retrieval representations without invoking the target Video-LLM's expensive visual encoder.
When a query arrives, WRWS estimates retrieval uncertainty from the query--observation similarity distribution and uses it to modulate the semantic--recency trade-off: concentrated similarity favors semantic relevance, whereas diffuse similarity strengthens the recency prior.
The resulting policy selects a bounded set of observations before invoking the target Video-LLM only on the selected evidence.
Because evidence selection is model-external, WRWS requires neither parameter updates nor access to target-model hidden states, attention statistics, or historical KV caches.

Evaluation spans four Video-LLM families and multiple model scales on streaming video understanding benchmarks~\citep{lin2024streaming,niu2025ovo}, with additional experiments on offline long-video understanding.
Across these settings, WRWS achieves competitive or improved online performance against strong training-free baselines.
Compared with SimpleStream~\citep{simplestream}, WRWS reduces VTAT to approximately $48\%$ while maintaining competitive downstream accuracy.
Its model-external design is also compatible with vLLM~\citep{kwon2023efficient}, demonstrating integration with an optimized multimodal serving backend.

Our contributions are threefold:
\begin{itemize}

    \item
    We formulate streaming Video-LLM inference as temporal evidence allocation under a bounded target-model visual-processing budget.
    Query-level analysis reveals complementarity between Recent-only and Semantic-only selection.
    WRWS therefore uses retrieval uncertainty to adaptively balance semantic relevance and temporal recency for each query.
        
    \item
    WRWS selects Top-$K$ evidence using lightweight external representations before invoking the target Video-LLM, thereby restricting expensive target-model visual processing to only the selected observations and keeping the per-query visual budget bounded.

    \item 
    Across multiple Video-LLM families and model scales, WRWS achieves competitive or improved streaming performance while reducing VTAT to approximately $48\%$ the state-of-the-art method under standard Transformers inference.

\end{itemize}

\section{Empirical Observations}
\label{sec:empirical_observations}

\begin{figure*}[t]
    \centering

    \begin{minipage}[t]{0.37\textwidth}
        \vspace{0pt}
        \centering

        \vspace{-2mm}

        \includegraphics[width=\linewidth]{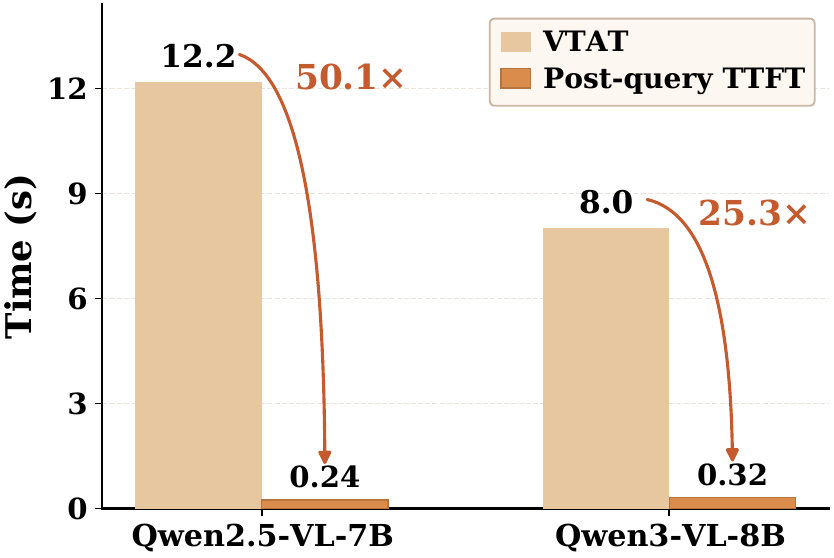}

        \vspace{-3mm}

        \captionof{figure}{
            \textbf{Profiling target-VLM visual-processing cost.}
            VTAT is $50.1\times$ and $25.3\times$ inference TTFT
            for Qwen2.5/3-VL.
        }
        \label{fig:observation1}
    \end{minipage}
    \hfill
    \begin{minipage}[t]{0.61\textwidth}
        \vspace{0pt}
        \centering

        \captionof{table}{
            \textbf{Fixed evidence policies leave substantial query-level headroom.}
            Recent-only outperforms Semantic-only under the same evidence budget,
            while a per-query oracle consistently improves over both.
        }
        \label{tab:observation2}

        \vspace{2mm}

        \scriptsize
        \renewcommand{\arraystretch}{1.2}
        \setlength{\tabcolsep}{4.5pt}

        \begin{tabular}{@{}ll|cccc@{}}
            \toprule
            \textbf{Backbone} & \textbf{Policy} & \textbf{Backward} & \textbf{Real-Time} & \textbf{Forward} & \textbf{Avg.} \\
            \midrule

            \multirow{3}{*}{Qwen2.5-VL-7B}
            & Semantic-only & 47.12 & 59.08 & 39.17 & 48.46 \\
            & Recent-only   & 50.43 & 77.20 & 39.56 & 55.73 \\
            & \cellcolor{gray!15}\textbf{Oracle} & \cellcolor{gray!15}\textbf{59.21} & \cellcolor{gray!15}\textbf{81.77} & \cellcolor{gray!15}\textbf{45.63} & \cellcolor{gray!15}\textbf{62.20} \\[1pt]

            \midrule

            \multirow{3}{*}{Qwen3-VL-8B}
            & Semantic-only & 48.46 & 62.87 & 43.95 & 51.76 \\
            & Recent-only   & 49.49 & 80.70 & 44.11 & 58.10 \\
            & \cellcolor{gray!15}\textbf{Oracle} & \cellcolor{gray!15}\textbf{60.56} & \cellcolor{gray!15}\textbf{84.45} & \cellcolor{gray!15}\textbf{52.25} & \cellcolor{gray!15}\textbf{65.75} \\

            \bottomrule
        \end{tabular}
    \end{minipage}

    \vspace{-5mm}
\end{figure*}

\subsection{Do We Need to Process Every Incoming Observation with the Target VLM?}

A straightforward streaming pipeline processes incoming observations with the target Video-LLM as they arrive, making their visual representations available when a query is issued.
However, this commits expensive computation before knowing which observations will ultimately matter for a future query.
We first quantify the cost of this encode-first strategy in practice.

Fig.~\ref{fig:observation1} compares vision-to-answer time (VTAT) with inference time-to-first-token (TTFT).
Both metrics end at the first generated answer token.
VTAT is measured from the first visual observation, whereas inference TTFT starts when the target Video-LLM begins inference for answering.
Under a fixed hardware and inference stack, VTAT therefore captures the cumulative runtime cost of processing the observed stream, while inference TTFT isolates the response cost after model inference.
For Qwen2.5-VL-7B and Qwen3-VL-8B, VTAT is $12.18$ s and $8.01$ s, respectively, compared with only $0.243$ s and $0.317$ s for inference TTFT.
Thus, VTAT is $50.1\times$ and $25.3\times$ the corresponding inference TTFT.
Under our profiling protocol, this large gap shows that most of the measured vision-to-answer time is spent outside the target-model inference stage.
A detailed VTAT breakdown in \textbf{Appendix}~\ref{app:empirical_analysis} further shows that target-VLM visual encoding is the largest component of this runtime.
This profiling result motivates treating target-model visual processing as a limited inference-time resource rather than applying it indiscriminately to every incoming observation.

\subsection{Is a Fixed Evidence Policy Sufficient?}

Once expensive target-model encoding is deferred until query time, the next question is how to allocate this limited computation. 
A simple strategy is to prioritize the most recent observations, which provides a strong temporal prior for streaming queries. 
However, a fixed recency policy may overlook earlier observations that are more relevant to the current query.

Tab.~\ref{tab:observation2} compares Recent-only with Semantic-only, which instead ranks observations by query-observation similarity under the same fixed evidence budget. 
Recent-only performs better on average for both Qwen2.5-VL-7B and Qwen3-VL-8B, reaching $55.73$ and $58.10$, versus $48.46$ and $51.76$ for Semantic-only, respectively. 
The advantage is especially pronounced on Real-Time tasks, where recent observations provide a particularly strong temporal prior.

The aggregate advantage of recency, however, does not mean that semantic retrieval is unnecessary. 
To examine their complementarity more directly, we construct a per-query oracle that selects the better result between Recent-only and Semantic-only for each query. 
The oracle reaches $62.20$ with Qwen2.5-VL-7B and $65.75$ with Qwen3-VL-8B, improving over Recent-only by $6.47$ and $7.65$ points, respectively. 
The gains are also consistent across all three task groups. 
The oracle gap reveals substantial query-level complementarity between the two fixed policies. This motivates WRWS to decide when to retrieve semantic evidence and when to retain recent context at query time.


\section{Method}
\subsection{Problem Setup}
Consider a video stream that produces a sequence of observations $\mathcal{H}_t=\{x_1,\ldots,x_t\}$ up to time $t$.
When a query $q_t$ arrives, a streaming Video-LLM must answer it using only the causally available history $\mathcal{H}_t$. 
Future observations $x_{t+1:}$ are unavailable.

Encoding every incoming observation with the target Video-LLM incurs expensive visual computation before the query's evidence needs are known.
We therefore impose a bounded visual-processing budget at query time. 
Given the available history $\mathcal{H}_t$, the system selects a subset $\mathcal{S}_t \subseteq \mathcal{H}_t$ with $|\mathcal{S}_t| \leq K$, where $K$ denotes the maximum number of observations that may receive expensive target-VLM visual processing for each query.
The selected observations in $\mathcal{S}_t$ are then encoded by the target Video-LLM and combined with $q_t$ for answer generation, while the remaining observations bypass target-model visual encoding for that query.

The resulting problem is to allocate the target-model visual-processing budget across the causally available history at query time.
Motivated by Sec.~\ref{sec:empirical_observations}, the selection policy should account for both semantic relevance and temporal recency rather than committing to either fixed evidence policy.
WRWS addresses this problem by using lightweight representations to retrieve candidate evidence and adapt the semantic--recency trade-off according to retrieval uncertainty.
It then selects $\mathcal{S}_t$ and invokes the target Video-LLM only on these selected observations for answer generation.

\begin{figure}[t]
    \centering
      \includegraphics[width=.99\linewidth]{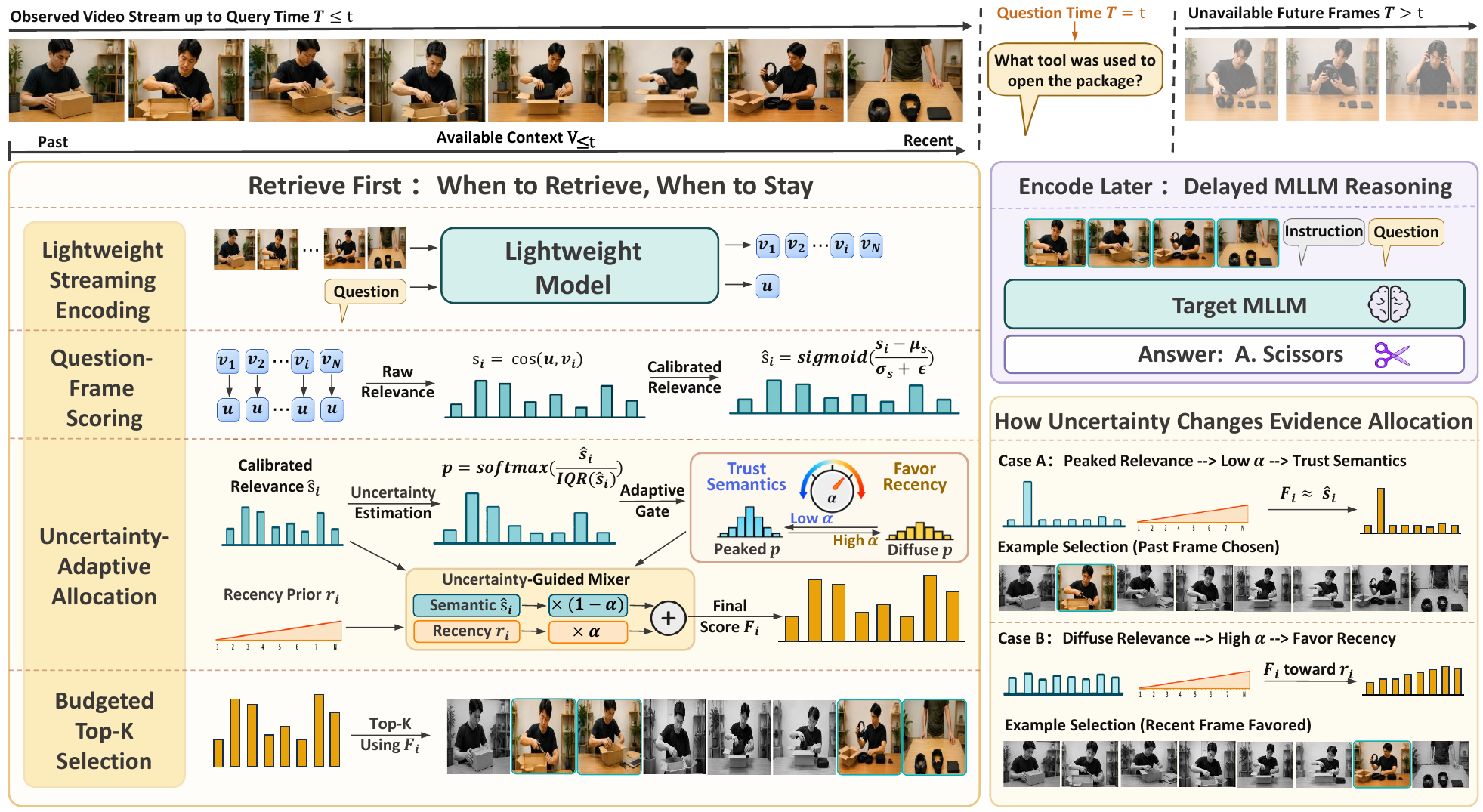}
      \vspace{-1mm}
      \caption{
      \textbf{Overview of WRWS.}
      WRWS uses lightweight representations to estimate query--observation relevance and retrieval uncertainty over the causally available history.
      The uncertainty adaptively balances semantic relevance and temporal recency to rank observations under a fixed Top-$K$ budget.
      Only the selected observations are then encoded by the target Video-LLM, realizing our \emph{retrieve-first, encode-later} strategy.
      The lower-right panel illustrates how peaked and diffuse relevance distributions shift evidence allocation toward semantics and recency, respectively.
      }
      \vspace{-3mm}
    \label{fig:method}
\end{figure}

\subsection{Lightweight Query--Observation Retrieval}

As illustrated in Fig.~\ref{fig:method}, WRWS first maintains lightweight representations of the observed stream, avoiding target Video-LLM invocation before evidence selection.
As each observation $x_i$ arrives, a lightweight vision-language encoder maps it to a visual representation $\mathbf{v}_i$, which is cached for subsequent retrieval.
When query $q_t$ arrives, the same model produces its text representation $\mathbf{u}$, and we compute the semantic relevance of each available observation by cosine similarity:
\begin{equation}
s_i = \operatorname{cos}(\mathbf{u}, \mathbf{v}_i),
\qquad x_i \in \mathcal{H}_t.
\end{equation}

Because similarity scores vary across queries, we calibrate them over the causally available history:
\begin{equation}
\hat{s}_i =
\operatorname{sigmoid}
\left(
\frac{s_i-\mu_s}{\sigma_s+\epsilon}
\right),
\end{equation}
where $\mu_s$ and $\sigma_s$ denote the mean and population standard deviation of $\{s_i\}_{x_i\in\mathcal{H}_t}$, and $\epsilon=10^{-6}$ is a small constant for numerical stability.
The calibrated score $\hat{s}_i$ provides a bounded semantic relevance signal for subsequent uncertainty estimation and adaptive semantic--recency allocation.

\subsection{Uncertainty-Aware Temporal Evidence Allocation}

\noindent\textbf{Retrieval Uncertainty.}
Semantic retrieval is reliable when relevance is concentrated on a few observations, but becomes ambiguous when many observations receive similar scores.
To quantify this ambiguity, we first convert the calibrated relevance scores into a probability distribution:
\begin{equation}
p_i =
\frac{
\exp\left((\hat{s}_i-m_t)/\tau_t\right)
}{
\sum_{j=1}^{N_t}
\exp\left((\hat{s}_j-m_t)/\tau_t\right)
},
\qquad
m_t=\max_j \hat{s}_j,
\end{equation}
where $\hat{\mathbf{s}}=\{\hat{s}_i\}_{i=1}^{N_t}$ and $N_t=|\mathcal{H}_t|$.
For $N_t>1$, the adaptive temperature is defined as
\begin{equation}
\tau_t =
\begin{cases}
\operatorname{IQR}(\hat{\mathbf{s}}),
& \operatorname{IQR}(\hat{\mathbf{s}}) > \epsilon, \\[1mm]
\operatorname{Std}(\hat{\mathbf{s}}),
& \operatorname{IQR}(\hat{\mathbf{s}}) \le \epsilon
\ \text{and}\
\operatorname{Std}(\hat{\mathbf{s}}) > \epsilon.
\end{cases}
\end{equation}
If both dispersion measures are at most $\epsilon$, the calibrated relevance scores are effectively indistinguishable and we use the uniform distribution $p_i=1/N_t$.
For $N_t=1$, we set $p_1=1$.
The adaptive temperature adjusts the softmax scale to the dispersion of calibrated relevance scores for each query.
We then use normalized entropy as the retrieval uncertainty:
\begin{equation}
\alpha_t =
\begin{cases}
\displaystyle
\frac{-\sum_{i=1}^{N_t} p_i \log p_i}
{\log N_t},
& N_t>1, \\[3mm]
0,
& N_t=1,
\end{cases}
\end{equation}
where $\alpha_t\in[0,1]$.
For a singleton history, no evidence-allocation decision is required, so we set $\alpha_t=0$ by convention.
As illustrated in the lower-right panel of Fig.~\ref{fig:method}, a peaked relevance distribution yields low $\alpha_t$ and places greater trust in semantic relevance, while a diffuse distribution yields high $\alpha_t$ and shifts the allocation toward recent observations.

\noindent\textbf{Recency Prior.}
Semantic relevance alone does not capture the temporal prior of streaming queries, which often favors observations closer to the query time.
We therefore assign each observation a recency score $r_i\in[0,1]$ that increases monotonically with its temporal proximity to the query:
\begin{equation}
r_i =
\begin{cases}
\displaystyle
\frac{i-1}{N_t-1},
& N_t>1, \\[2mm]
0,
& N_t=1,
\end{cases}
\qquad i=1,\ldots,N_t.
\end{equation}
Thus, more recent observations receive larger temporal weights, while earlier observations remain available for selection when supported by sufficiently strong semantic relevance.

\noindent\textbf{Adaptive Semantic--Recency Allocation.}
We use the retrieval uncertainty $\alpha_t$ to adaptively combine semantic relevance and temporal recency:
\begin{equation}
F_i =
(1-\alpha_t)\hat{s}_i
+
\alpha_t r_i.
\end{equation}
When retrieval is confident, a small $\alpha_t$ makes $F_i$ primarily follow semantic relevance.
As retrieval becomes more ambiguous, $\alpha_t$ increases and shifts the ranking toward recent observations.
Through $\alpha_t$, the score continuously interpolates between semantic relevance and temporal recency for each query in response to the estimated retrieval uncertainty.

\subsection{Budgeted Selection and Delayed Target-VLM Encoding}
Given the evidence scores $\{F_i\}_{i=1}^{N_t}$, we set $K_t=\min(K,N_t)$ and select the Top-$K_t$ observations from the causally available history for subsequent target-model processing:
\begin{equation}
\mathcal{I}_t
=
\operatorname{TopK_t}_{i\in\{1,\ldots,N_t\}} F_i,
\qquad
\mathcal{S}_t
=
\{x_i \mid i\in\mathcal{I}_t\}.
\end{equation}
When $N_t\le K$, all causally available observations are therefore selected.
Only the selected observations in $\mathcal{S}_t$ are passed to the target Video-LLM with $q_t$ for answering, while unselected observations bypass expensive target-model visual processing for the current query.

Evidence selection is therefore completed before query-specific target-VLM visual processing. 
The target model processes at most \(K\) observations per query, and no target-model states are required for retrieval.
This model-external design keeps target-model inference unchanged and therefore \textbf{supports optimized serving backends such as vLLM.}

\begin{table*}[t]

\centering

\caption{
Performance comparison (\%) on StreamingBench and OVO-Bench.
Methods prefixed with ``+'' are applied to the corresponding base model.
$\dagger$ denotes results reproduced following the official benchmark protocol.
Gray rows denote WRWS.
Best results are shown in \textbf{bold}.
}

\vspace{-2mm}

\label{tab:main_results_online}

\scriptsize

\setlength{\tabcolsep}{4.9pt}

\renewcommand{\arraystretch}{0.95}
\setlength{\aboverulesep}{0.25ex}
\setlength{\belowrulesep}{0.25ex}

\newcommand{\methodplus}{\hspace{0.8em}+\,}

\begin{tabular}{l|c|c|cccc}

\toprule

\multicolumn{1}{c|}{\multirow{2}{*}{\textbf{Model}}} &
\multicolumn{1}{c|}{\multirow{2}{*}{\textbf{\#Frames}}} &
\multicolumn{1}{|c|}{\textbf{StreamingBench}} &
\multicolumn{4}{c}{\textbf{OVO-Bench}} \\
\hhline{~~|-|----}
& &
\textbf{Real-Time} &
\textbf{Backward} &
\textbf{Real-Time} &
\textbf{Forward} &
\textbf{Avg.} \\

\midrule
Human & -- & 91.46 & 92.33 & 93.20 & 92.90 & 92.81 \\
\midrule
\multicolumn{6}{c}{\textit{Proprietary MLLMs}} \\
\midrule
Gemini 1.5 Pro~\citep{gemini15}  &  1 fps
& 75.69 & 62.54 & 69.32 & 57.15 & 63.00 \\
GPT-4o~\citep{gpt4o} &  64 
& 73.28 & 60.75 & 64.46 & 53.40 & 59.54 \\
Claude 3.5 Sonnet~\citep{claude35} & 20
& 72.44 & -- & -- & -- & -- \\
\midrule
\multicolumn{6}{c}{\textit{Open-source Offline MLLMs}} \\
\midrule
LLaVA-Video-7B~\citep{llavavideo} & 64
& -- & 40.40 & 63.52 & 54.82 & 52.91 \\
Qwen2-VL-7B~\citep{qwen2vl}  & 64
& 69.04 & 46.46 & 55.98 & 48.74 & 50.39 \\
InternVL2-8B~\citep{internvl2} & 16
& 63.72 & 43.44 & 60.39 & 46.60 & 50.15 \\
LongVU-7B~\citep{longvu} & 1 fps
& -- & 35.01 & 57.61 & 47.50 & 46.71 \\
\midrule
\multicolumn{6}{c}{\textit{Open-source Online / Streaming MLLMs}} \\
\midrule

VideoLLM-online-8B~\citep{videollm_online} & 2 fps
& 35.99 & 17.73 & 20.79 & -- & -- \\
Flash-VStream-7B~\citep{flashvstream} & 1 fps
& 23.23 & 27.38 & 28.37 & 45.09 & 33.61 \\
Dispider-7B~\citep{dispider} & 1 fps
& 67.63 & 36.06 & 54.55 & 34.72 & 41.78 \\
TimeChat-Online-7B~\citep{timechatonline} & 1 fps
& 75.28 & 44.50 & 61.40 & 36.80 & 47.60 \\
StreamForest-7B~\citep{streamforest} & 1 fps
& 77.26 & 52.02 & 61.20 & 53.49 & 55.57 \\
Streamo-7B (1 fps)~\citep{streamo} & 1 fps
& -- & 46.10 & 65.98 & 54.77 & 55.61 \\
\midrule
\multicolumn{6}{c}{\textit{Training-free Offline-to-Online Methods}} \\
\midrule
LLaVA-OV-0.5B~\citep{llavaonevision} & 64
& 59.64 & 34.59 & 49.70 & 41.08 & 41.79 \\
\methodplus ReKV~\citep{rekv} & 0.5 fps
& 57.39 & 33.06 & 43.77 & -- & -- \\
\methodplus HERMES (6K tokens)~\citep{hermes} & 0.5 fps
& 61.04 & 34.75 & 50.34 & -- & -- \\
\methodplus HERMES (4K tokens)~\citep{hermes} & 0.5 fps
& \textbf{62.04} & 34.80 & 50.72 & -- & -- \\
\rowcolor{gray!18}
\methodplus WRWS (Ours) & 0.5 fps
& 61.54 & \textbf{38.57} & \textbf{53.12} & 41.09 & 44.26 \\[1pt]
\midrule
LLaVA-OV-7B~\citep{llavaonevision} & 64
& 71.34 & 43.71 & 64.02 & 50.50 & 52.74 \\
\methodplus ReKV~\citep{rekv} & 0.5 fps
& 69.22 & 44.16 & 57.33 & -- & -- \\
\methodplus HERMES (6K tokens)~\citep{hermes} & 0.5 fps
& 72.63 & 48.80 & 65.07 & -- & -- \\
\methodplus HERMES (4K tokens)~\citep{hermes} & 0.5 fps
& 73.23 & 50.20 & 66.34 & -- & -- \\
\rowcolor{gray!18}
\methodplus WRWS (Ours) & 0.5 fps
& \textbf{75.35} & \textbf{50.78} & \textbf{67.45} & \textbf{51.09} & \textbf{56.44} \\
\midrule
Qwen2.5-VL-7B~\citep{qwen25vl} & 1 fps
& 73.31 & 44.65 & 59.90 & 39.02 & 47.86 \\
\methodplus HERMES (6K tokens)~\citep{hermes} & 1 fps
& 78.72 & 48.10 & 68.42 & -- & -- \\
\methodplus HERMES (4K tokens)~\citep{hermes} & 1 fps
& 79.44 & 49.43 & 68.98 & -- & -- \\
\methodplus FluxMem~\citep{FluxMem} & 1 fps
& 76.40 & 47.24 & 67.20 & -- & -- \\
\methodplus SAVEMem~\citep{SAVEMem} & 1 fps
& 76.00 & 50.44 & 74.93 & -- & -- \\
\methodplus SimpleStream$\dagger$~\citep{simplestream} & 1 fps
& 78.47 & 50.43 & \textbf{77.20} & 39.56 & 55.73 \\
\rowcolor{gray!18}
\methodplus WRWS (Ours) & 1 fps
& \textbf{79.47} & \textbf{50.59} & 77.13 & \textbf{40.23} & \textbf{55.98} \\[1pt]
\midrule
Qwen2.5-VL-32B~\citep{qwen25vl} & 1 fps
& 74.27 & 50.33 & 64.40 & -- & -- \\
\methodplus HERMES (6K tokens)~\citep{hermes} & 1 fps
& 80.20 & \textbf{57.71} & 71.93 & -- & -- \\
\methodplus HERMES (4K tokens)~\citep{hermes} & 1 fps
& 80.08 & 55.42 & 72.37 & -- & -- \\
\methodplus SimpleStream$\dagger$~\citep{simplestream} & 1 fps
& \textbf{80.31} & 49.03 & \textbf{80.58} & 42.05 & \textbf{57.22} \\
\rowcolor{gray!18}
\methodplus WRWS (Ours) & 1 fps
& 80.23 & 49.28 & 80.13 & \textbf{42.25} & \textbf{57.22} \\[1pt]
\midrule
Qwen3-VL-8B~\citep{qwen3vl} & 1 fps
& 80.51 & 44.00 & 65.00 & 46.30 & 51.77 \\
\methodplus SimpleStream$\dagger$~\citep{simplestream} & 1 fps
& 80.59 & 48.84 & \textbf{81.26} & 42.55 & 57.55 \\
\rowcolor{gray!18}
\methodplus WRWS (Ours) & 1 fps
& \textbf{81.11} & \textbf{50.00} & 80.74 & \textbf{45.05} & \textbf{58.59} \\[1pt]
\midrule
Qwen3-VL-30B-A3B~\citep{qwen3vl} & 1 fps
& 80.75 & 63.24 & 75.43 & 59.15 & 65.94 \\
\methodplus SimpleStream$\dagger$~\citep{simplestream} & 1 fps
& 83.51 & 61.45 & \textbf{85.48} & 44.12 & 63.68 \\
\rowcolor{gray!18}
\methodplus WRWS (Ours) & 1 fps
& \textbf{83.71} & \textbf{62.35} & 84.94 & \textbf{45.77} & \textbf{64.35} \\[1pt]
\midrule
Qwen3.5-4B~\citep{qwen35} & 1 fps
& 77.03 & 56.61 & 67.99 & \textbf{52.49} & 59.03 \\
\rowcolor{gray!18}
\methodplus WRWS (Ours) & 1 fps
& \textbf{78.39} & \textbf{57.00} & \textbf{79.83} & 45.24 & \textbf{60.69} \\[1pt]
\midrule
Qwen3.5-35B-A3B~\citep{qwen35} & 1 fps
& \textbf{84.55} & 62.56 & 72.89 & \textbf{58.38} & 64.61 \\
\rowcolor{gray!18}
\methodplus WRWS (Ours) & 1 fps
& 82.99 & \textbf{65.07} & \textbf{82.79} & 46.40 & \textbf{64.75} \\[1pt]
\bottomrule
\end{tabular}
\vspace{-5mm}
\end{table*}

\begin{figure*}[t]
\centering

\begin{minipage}[t]{0.60\textwidth}
    \centering
    \vspace{0pt}

    \makeatletter
    \def\@captype{table}
    \makeatother

    \caption{
        Performance comparison (\%) on offline benchmarks.
        Methods prefixed with ``+'' are applied to the corresponding base model.
        Gray rows denote WRWS.
    }
    \vspace{-1mm}
    \label{tab:main_results_offline}

    \scriptsize
    \setlength{\tabcolsep}{3.5pt}
    \renewcommand{\arraystretch}{0.90}

    \setlength{\aboverulesep}{0.65ex}
    \setlength{\belowrulesep}{0.65ex}

    \newcommand{\methodplus}{\hspace{0.8em}+\,}

    \resizebox{\linewidth}{!}{
    \begin{tabular}{l|cccc}
        \toprule
        \multirow{2}{*}{\textbf{Model}} &
        \multirow{2}{*}{\textbf{MLVU}} &
        \multirow{2}{*}{\textbf{EgoSchema}} &
        \multicolumn{2}{c}{\textbf{VideoMME}} \\
        & & & \textbf{Long} & \textbf{Avg.} \\
        \midrule

        \multicolumn{5}{c}{\textit{Proprietary MLLMs}} \\
        \midrule

        Gemini 1.5 Pro~\citep{gemini15}
        & -- & 69.32 & 62.54 & 66.41 \\

        GPT-4o~\citep{gpt4o}
        & 64.60 & 64.46 & 60.75 & 62.87 \\

        \midrule
        \multicolumn{5}{c}{\textit{Open-source Offline MLLMs}} \\
        \midrule

        LLaVA-Video-7B~\citep{llavavideo}
        & 70.80 & 57.30 & -- & 63.30 \\

        Qwen2-VL-7B~\citep{qwen2vl}
        & -- & 66.70 & -- & 63.30 \\

        \midrule
        \multicolumn{5}{c}{\textit{Open-source Online MLLMs}} \\
        \midrule

        Dispider-7B~\citep{dispider}
        & 61.70 & 55.60 & -- & 57.20 \\

        TimeChat-Online-7B~\citep{timechatonline}
        & 62.60 & 61.90 & 41.70 & 53.22 \\

        StreamForest-7B~\citep{streamforest}
        & 70.00 & -- & -- & 61.40 \\

        \midrule
        \multicolumn{5}{c}{\textit{Training-free Offline-to-Online Methods}} \\
        \midrule

        LLaVA-OV-7B~\citep{llavaonevision}
        & 64.70 & 59.93 & 48.00 & 57.67 \\

        \methodplus ReKV~\citep{rekv}
        & 68.50 & \textbf{60.70} & 46.89 & 57.74 \\

        \methodplus HERMES (6K tokens)~\citep{hermes}
        & -- & 60.23 & \textbf{49.11} & 58.44 \\

        \rowcolor{gray!18}
        \methodplus WRWS (Ours)
        & \textbf{71.96} & 60.48 & 47.90 & \textbf{58.85} \\

        \bottomrule
    \end{tabular}
    }
\end{minipage}
\hspace{0.025\textwidth}
%
\begin{minipage}[t]{0.36\textwidth}
    \centering
    \vspace{-1mm}

    \makeatletter
    \def\@captype{figure}
    \makeatother

    \includegraphics[
        width=0.96\linewidth
    ]{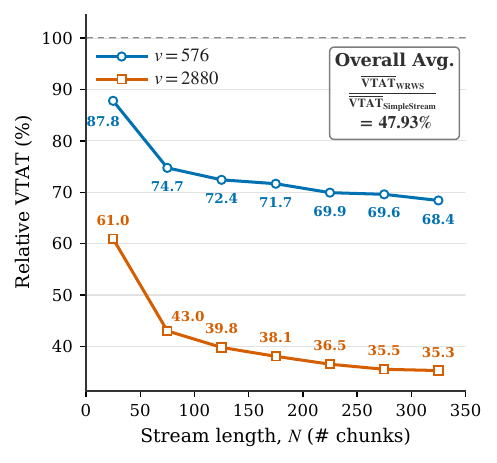}


    \begin{minipage}{0.99\linewidth}
        \centering
        \caption{
        Relative VTAT across stream lengths for two common visual-token loads covering over 60\% of samples. 
        47.93\% is averaged over all samples.
        }
        \label{fig:efficiency}
    \end{minipage}
\end{minipage}

\vspace{-6mm}
\end{figure*}

\section{Experiments}
\subsection{Experimental Setup}
\noindent\textbf{Benchmarks.}
We evaluate WRWS on StreamingBench~\citep{lin2024streaming} and OVO-Bench~\citep{niu2025ovo} for streaming video understanding, using all 12 OVO-Bench tasks across the Backward, Real-Time, and Forward groups.
We additionally evaluate on EgoSchema~\citep{mangalam2023egoschema}, Video-MME~\citep{fu2025videomme}, and MLVU~\citep{zhou2025mlvu} for offline video understanding.
We follow the official evaluation protocols unless otherwise specified.

\noindent\textbf{Implementation.}
WRWS uses SigLIP~\citep{zhai2023siglip} for lightweight retrieval.
Streaming inference is strictly causal and never accesses observations beyond the current streaming timestamp.
Full implementation and evaluation details are provided in \textbf{Appendix}~\ref{app:implementation}.

\subsection{Main Results}
\label{sec:main_results}

\noindent\textbf{Online benchmarks.}
Tab.~\ref{tab:main_results_online} compares WRWS with existing methods on StreamingBench and OVO-Bench.
Across the four Qwen2.5/3 backbones, WRWS matches or improves the OVO-Bench average over SimpleStream, with gains of up to 1.04 points, while remaining comparable on StreamingBench.
On LLaVA-OV-7B, WRWS achieves 75.35 on StreamingBench and 56.44 on OVO-Bench, outperforming the reported HERMES variants.
WRWS further extends to the recent Qwen3.5 family, improving the OVO-Bench average over processing all sampled observations on both the 4B and 35B-A3B models.
Results against HERMES under its 2-fps setting are provided in \textbf{Appendix}~\ref{app:hermes_2fps}.

\noindent\textbf{Across backbones and scales.}
WRWS applies without additional training across LLaVA-OV, Qwen2.5-VL, Qwen3-VL, and Qwen3.5, spanning models from 0.5B to 35B-A3B. 
This suggests that its evidence-allocation mechanism is not tied to a specific Video-LLM family or model scale. 
Additional results across more model sizes and evidence budgets are provided in \textbf{Appendix}~\ref{app:additional_online}.

\noindent\textbf{Offline benchmarks.}
Tab.~\ref{tab:main_results_offline} further evaluates WRWS on offline video benchmarks.
With LLaVA-OV-7B, WRWS achieves 71.96 on MLVU and a VideoMME average of 58.85, the highest among the compared training-free methods, while remaining competitive on EgoSchema and VideoMME.

\subsection{Efficiency}
\label{sec:efficiency}

Fig.~\ref{fig:efficiency} compares the VTAT of WRWS with SimpleStream on Qwen2.5-VL-7B as stream length increases.
We visualize two common per-observation visual-token loads, which together account for over 60\% of evaluation samples.
Across all samples, WRWS requires only 47.93\% of SimpleStream's VTAT on average.
The relative VTAT further decreases with stream length for both loads, with larger savings under the higher visual-token load.
This scaling behavior follows from avoiding expensive target-model visual processing over the full observed history as the stream grows.
Additional TTFT and GPU-memory results are provided in \textbf{Appendix}~\ref{app:efficiency}.

\begin{figure}[t]
    \centering
      \includegraphics[width=.99\linewidth]{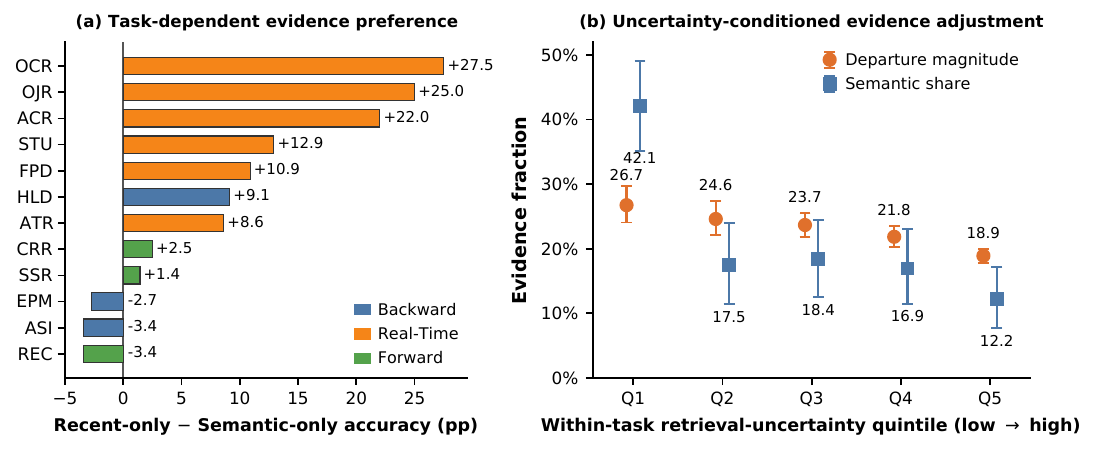}
      \vspace{-3mm}
      \caption{
      \textbf{Understanding adaptive evidence allocation.}
      \textbf{(a)} The relative utility of Recent-only and Semantic-only varies across temporal tasks.
      \textbf{(b)} When WRWS departs from Recent-only, lower retrieval uncertainty is associated with larger adjustments and a higher semantic share among the introduced evidence.
      Error bars denote 95\% within-task stratified bootstrap confidence intervals.
      }
      \vspace{-4mm}
    \label{fig:adaptive_behavior}
\end{figure}

\section{Understanding Adaptive Evidence Allocation}
\label{sec:adaptive_behavior}

Building on the observed complementarity between fixed evidence policies (Sec. \ref{sec:empirical_observations}), we examine how their relative utility varies across tasks and how WRWS resolves the trade-off at the query level.

\subsection{Task-level Behavior}

Fig.~\ref{fig:adaptive_behavior}(a) compares Recent-only and Semantic-only across the 12 OVO-Bench tasks.
Their relative utility varies across tasks.
Recent-only performs better on all six Real-Time Visual Perception tasks, with gains of $8.6$--$27.5$ percentage points, consistent with their reliance on current or nearby visual evidence.
In contrast, Semantic-only performs better on EPM, ASI, and REC, which can require earlier or temporally distributed evidence.
The preference is not determined solely by the benchmark-level task group: HLD favors Recent-only, while CRR and SSR show only modest differences.
This task-level variation supports adaptive semantic-recency allocation at query time.

\subsection{Sample-level Behavior}

We next examine how retrieval uncertainty affects WRWS's final evidence selection.
Let $\mathcal{A}_q$, $\mathcal{R}_q$, and $\mathcal{M}_q$ denote the evidence selected by WRWS, Recent-only, and Semantic-only for query $q$.
WRWS retains a strong recency prior, matching Recent-only exactly on $79.9\%$ of queries.
For the remaining queries, we measure the departure magnitude
\begin{equation}
D_q =
\frac{
    |\mathcal{A}_q \setminus \mathcal{R}_q|
}{
    |\mathcal{A}_q|
},
\end{equation}
which measures the extent of deviation from Recent-only, and the semantic replacement share
\begin{equation}
G_q =
\frac{
    |(\mathcal{A}_q \setminus \mathcal{R}_q)
    \cap
    (\mathcal{M}_q \setminus \mathcal{R}_q)|
}{
    |\mathcal{A}_q \setminus \mathcal{R}_q|
},
\end{equation}
which measures the semantic alignment of the introduced evidence.
Queries are grouped into retrieval-uncertainty quintiles independently within each task.

As shown in Fig.~\ref{fig:adaptive_behavior}(b), the average departure magnitude decreases from $26.7\%$ to $18.9\%$ from the lowest- to highest-uncertainty quintile, while the semantic replacement share decreases overall from $42.1\%$ to $12.2\%$.
Thus, WRWS maintains a strong recency prior while making larger and more semantically aligned adjustments when retrieval is more confident.

\begin{table*}[t]
\centering

\caption{
\textbf{Ablation studies of WRWS.}
$\dagger$ denotes the default setting. Best results are shown in \textbf{bold}.
}
\label{tab:ablation}
\vspace{-3mm}

\scriptsize

\renewcommand{\arraystretch}{0.95}
\setlength{\aboverulesep}{0.25ex}
\setlength{\belowrulesep}{0.25ex}


\noindent
\begin{minipage}[t]{0.485\textwidth}
\centering

\textbf{(a) Adaptive evidence allocation}

{\renewcommand{\arraystretch}{0.99}
\begin{tabular*}{\linewidth}{@{\extracolsep{\fill}}lcc@{}}
\toprule
Allocation & Qwen2.5 & Qwen3 \\
\midrule
$\alpha=0.00$ (Semantic-only) & 48.46 & 51.76 \\
$\alpha=0.50$                 & 52.84 & 56.15 \\
$\alpha=1.00$ (Recent-only)   & 55.73 & 58.10 \\
Adaptive$^\dagger$            & \textbf{55.98} & \textbf{58.59} \\
\bottomrule
\end{tabular*}
}

\end{minipage}%
\hfill%
\begin{minipage}[t]{0.485\textwidth}
\centering

\textbf{(b) Recency prior}

{\renewcommand{\arraystretch}{0.99}
\begin{tabular*}{\linewidth}{@{\extracolsep{\fill}}lccc@{}}
\toprule
& Linear$^\dagger$ & Plain & Exponential \\
\midrule
Back. & \textbf{50.00} & 48.46 & 49.71 \\
RT    & \textbf{80.74} & 62.87 & 80.73 \\
Fwd.  & \textbf{45.05} & 43.95 & 44.48 \\
All   & \textbf{58.59} & 51.76 & 58.31 \\
\bottomrule
\end{tabular*}
}

\end{minipage}

\par


\noindent
\begin{minipage}[t]{0.485\textwidth}
\centering

\textbf{(c) Retrieval uncertainty}

{\renewcommand{\arraystretch}{0.99}
\begin{tabular*}{\linewidth}{@{\extracolsep{\fill}}lcccc@{}}
\toprule
& Entropy$^\dagger$ & Margin & Max & Std \\
\midrule
Back. & \textbf{50.00} & 49.49 & 49.49 & 49.94 \\
RT    & \textbf{80.74} & 80.70 & 80.59 & 80.36 \\
Fwd.  & 45.05 & 44.11 & 44.11 & \textbf{45.39} \\
All   & \textbf{58.59} & 58.10 & 58.06 & 58.56 \\
\bottomrule
\end{tabular*}
}

\end{minipage}
\hfill
\begin{minipage}[t]{0.485\textwidth}
\centering

\textbf{(d) Retrieval encoder}

{%
\renewcommand{\arraystretch}{0.99}
\setlength{\tabcolsep}{3.0pt}

\begin{tabular*}{\linewidth}{@{\extracolsep{\fill}}lcccc@{}}
\toprule
& SigLIP$^\dagger$ & CLIP-L & SigLIP2 & Target-VLM \\
\midrule
Back. & \textbf{50.00} & 49.37 & 49.59 & 48.59 \\
RT    & 80.74 & \textbf{80.91} & 80.71 & 80.01 \\
Fwd.  & \textbf{45.05} & 45.04 & 44.90 & 44.08 \\
All   & \textbf{58.59} & 58.44 & 58.40 & 57.56 \\
\bottomrule
\end{tabular*}
}

\end{minipage}

\vspace{-5mm}

\end{table*}

\section{Ablation Studies}


\noindent\textbf{Adaptive allocation.}
To assess the benefit of adaptive allocation, we compare it with representative fixed $\alpha$ values on Qwen2.5-VL and Qwen3-VL.
As shown in Tab.~\ref{tab:ablation}(a), adaptive allocation outperforms the fixed settings reported in the main table on both backbones.
The detailed $\alpha$ sweep in \textbf{Appendix}~\ref{app:additional_ablation} further shows that the best fixed setting is backbone-dependent: $\alpha=1.0$ for Qwen2.5-VL and $\alpha=0.75$ for Qwen3-VL.
Adaptive allocation removes the need to select such a backbone-specific trade-off and outperforms the best fixed setting by 0.25 and 0.34 points, respectively.

\noindent\textbf{Recency prior.}
Tab.~\ref{tab:ablation}(b) compares linear and exponential recency weighting with a variant that removes it.
Removing the recency prior reduces the score from 58.59 to 51.76, whereas linear and exponential weighting differ by only 0.28 points, suggesting that the presence of a recency prior matters more than its exact functional form.
We use the simpler linear weighting by default.

\noindent\textbf{Retrieval uncertainty.}
For retrieval uncertainty, we consider entropy, maximum confidence, margin, and standard deviation.
As shown in Tab.~\ref{tab:ablation}(c), entropy and standard deviation, which characterize the similarity distribution as a whole, outperform maximum confidence and margin, which depend only on the top-ranked scores.
We use entropy by default because it directly measures distributional uncertainty and achieves the best overall result.

\noindent\textbf{Lightweight retrieval encoding.}
Tab.~\ref{tab:ablation}(d) compares three lightweight retrieval encoders with representations from the target Video-LLM.
The lightweight encoders perform similarly, with SigLIP achieving the highest score of 58.59.
Using target-model representations lowers the score to 57.56 despite requiring the expensive visual encoder of the Video-LLM.
These results further support the use of lightweight external representations for retrieval before target-model visual processing.

\section{Related Work}

\noindent\textbf{Streaming Video Understanding with Video-LLMs.}
Existing approaches enable streaming video understanding through streaming-specific training and architectures~\citep{flashvstream,videochat_online,videollm_online,videostreaming}, or through training-free management of historical representations and redundant computation~\citep{streamchat,rekv,hermes,mukv,stc,streamingtom}.
The latter organize or compress memory and KV states, filter observations, or reuse visual representations to control context growth and computation.
These methods optimize how observations are represented, retained, or reused within the target-model processing pipeline.
WRWS focuses on an earlier decision: which causally available observations should receive expensive target-model visual processing in the first place.

\noindent\textbf{Training-Free Temporal Evidence Selection.}
SimpleStream~\citep{simplestream} establishes recent-frame retention as a strong training-free baseline, but allocates evidence solely by temporal proximity.
Other methods adapt access to historical evidence: OASIS~\citep{oasis} uses preliminary MLLM reasoning over recent context and event summaries to trigger retrieval, while ShallowStream~\citep{shallowstream} uses target-model logits, KV states, and attention for routing and evidence selection.
In contrast, WRWS adapts the semantic--recency trade-off before target-model visual encoding using retrieval uncertainty estimated from lightweight external representations.
Its selection policy is fully model-external, requiring neither preliminary MLLM reasoning nor access to target-model attention or historical KV states.

\section{Conclusion}
WRWS addresses a central question in streaming Video-LLM inference: when should the model retrieve semantically relevant history, and when should it stay with recent context?
Our experiments show that this choice varies across queries.
Using retrieval uncertainty to make the choice allows WRWS to select evidence before target-model visual encoding, reducing average VTAT to 47.93\% of SimpleStream while maintaining comparable downstream performance. 
The same formulation applies across multiple Video-LLM families and model scales without additional training.

\bibliographystyle{abbrvnat}
\bibliography{report}

\clearpage
\appendix

\section{Overview}

The supplementary material is organized as follows:

\begin{itemize}

    \item \textbf{Detailed Experimental Setup (Sec.~\ref{app:implementation}).}
    We provide details on the target Video-LLMs, frame-sampling rates, retrieval encoder, evidence budgets, inference configuration, hardware, and latency measurement protocol.

    \item \textbf{Additional Analysis of Empirical Observations (Sec.~\ref{app:empirical_analysis}).}
    We provide a detailed VTAT breakdown and per-task results for Semantic-only, Recent-only, and the per-query oracle.

    \item \textbf{Additional Online Evaluation (Sec.~\ref{app:additional_online}).}
    We report per-task results, comparisons with HERMES under the 2-fps setting, performance under different visual-processing budgets, and results across additional model scales.

    \item \textbf{Additional Analysis of Adaptive Evidence Allocation (Sec.~\ref{app:additional_ablation}).}
    We provide detailed comparisons of adaptive and fixed evidence allocation, analyze how the preferred semantic--recency trade-off varies across models and tasks, and examine how retrieval uncertainty relates to the relative utility of the two evidence policies.

    \item \textbf{Additional Efficiency Analysis (Sec.~\ref{app:efficiency}).}
    We provide detailed comparisons of VTAT, inference TTFT, and GPU memory usage.

    \item \textbf{Qualitative Analysis and Failure Cases (Sec.~\ref{app:qualitative}).}
    We present representative examples of semantic--recency allocation and analyze failures arising from evidence selection and downstream Video-LLM prediction.

    \item \textbf{Limitations and Future Work (Sec.~\ref{app:limitations}).}
    We discuss retrieval-cache growth with stream length and the dependence of adaptive allocation on uncertainty estimated from external retrieval representations.

\end{itemize}

\section{Detailed Experimental Setup}
\label{app:implementation}

\noindent\textbf{Target Video-LLMs and frame sampling.}
We evaluate WRWS with eight target Video-LLMs: LLaVA-OneVision-0.5B and 7B, Qwen2.5-VL-7B and 32B, Qwen3-VL-8B and 30B-A3B, and Qwen3.5-4B and 35B-A3B.
For streaming evaluation, we sample video observations at 0.5 fps for LLaVA-OneVision and 1 fps for the Qwen2.5-VL, Qwen3-VL, and Qwen3.5 families.
For each target model, we use the same sampling rate across StreamingBench and OVO-Bench.

\noindent\textbf{Retrieval and evidence budget.}
We use \texttt{google/siglip-so400m-patch14-384}~\citep{zhai2023siglip} as the lightweight retrieval encoder.
For streaming benchmarks, we set the evidence budget $K$ to 16 for LLaVA-OneVision, 4 for Qwen2.5-VL, and 6 for Qwen3-VL and Qwen3.5.
These family-specific budgets keep the maximum number of visual tokens processed by the target Video-LLM at approximately 3K.
We use the linear recency prior for all streaming experiments.
The adaptive semantic--recency weighting introduces no additional tuned hyperparameters.

\noindent\textbf{Offline evaluation.}
We evaluate EgoSchema, Video-MME, and MLVU with LLaVA-OneVision using the same sequential processing pipeline as in the streaming setting.
We sample video observations at 0.5 fps and set $K=32$, corresponding to approximately 6K target-model visual tokens and matching the 6K budget used by HERMES~\citep{hermes}.
Since these benchmarks do not provide the same streaming-time recency semantics as online evaluation, we disable the recency prior by assigning an identical constant temporal weight to all candidate observations, making the temporal term effectively uniform.
All other WRWS components remain unchanged.

\noindent\textbf{Inference configuration.}
Our implementation builds on the SimpleStream codebase~\citep{simplestream}.
We retain its target-model preprocessing and generation protocol, using the native processor of each Video-LLM and greedy decoding with at most 256 newly generated tokens.
For streaming benchmarks, inference is strictly causal: at any streaming timestamp, WRWS only accesses observations that have arrived by that timestamp.
All accuracy and efficiency results are obtained using standard Hugging Face Transformers inference.

\noindent\textbf{Hardware and latency measurement.}
Experiments are conducted on two NVIDIA A100 80GB GPUs, while latency is measured on NVIDIA 4090 GPUs.
We report two latency measures.
Vision-to-answer time (VTAT) measures the elapsed time from the start of processing the first video observation to the generation of the first answer token, whereas inference time-to-first-token (TTFT) measures the elapsed time from the start of target Video-LLM inference for answering to the generation of the first answer token.
We report the average VTAT and TTFT over all samples in OVO-Bench.

\noindent\textbf{vLLM deployment.}
We additionally implement WRWS with vLLM to verify its compatibility with an optimized multimodal serving backend.
The vLLM implementation is used only for the deployment study, with all other model configurations and WRWS settings kept identical to the standard setup in the main experiments.
\textbf{All accuracy and efficiency results in the main experiments are obtained with the Transformers implementation.}

\begin{table}[t]
    \centering
    \caption{
        \textbf{Breakdown of VTAT for Qwen2.5-VL-7B under the encode-first baseline.}
        Target-VLM visual encoding accounts for 62.36\% of total VTAT.
        Other pipeline overhead is the residual after subtracting target-VLM visual encoding and inference TTFT from VTAT, which includes processing not captured by those two timers.
    }
    \label{tab:vtat_breakdown}
    \vspace{-2mm}

    {
    \footnotesize
    \renewcommand{\arraystretch}{1.12}
    \setlength{\tabcolsep}{3.5pt}

    \begin{tabularx}{0.92\columnwidth}{lCCC|C}
        \toprule
        Metric
        & \makecell{Target-VLM\\visual encoding}
        & \makecell{Other pipeline\\overhead}
        & \makecell{Inference\\TTFT}
        & \makecell{Total\\VTAT} \\
        \midrule
        Time (s) & \textbf{7.596} & 4.342 & 0.243 & 12.181 \\
        Fraction (\%) & \textbf{62.36} & 35.65 & 1.99 & 100.00 \\
        \bottomrule
    \end{tabularx}
    }
\end{table}

\begin{table*}[t]
    \centering
    \caption{
        \textbf{Per-task breakdown of fixed evidence policies and the per-query oracle on OVO-Bench.}
        Semantic-only and Recent-only use the same fixed evidence budget.
        Oracle selects the better result between the two policies for each query.
        Gray rows denote the Oracle.
    }
    \label{tab:oracle_tasks}
    \vspace{-2mm}

    {
    \footnotesize
    \renewcommand{\arraystretch}{1.35}
    \setlength{\tabcolsep}{2.8pt}
    \setlength{\extrarowheight}{0.5pt}

    \begin{adjustbox}{max width=\textwidth}
    \begin{tabular}{c|l|ccc|cccccc|ccc}
        \hline
        \multirow{2}{*}{Model} & \multirow{2}{*}{Policy} & \multicolumn{3}{c|}{Backward} & \multicolumn{6}{c|}{Real-Time} & \multicolumn{3}{c}{Forward} \\
        \cline{3-5}
        \cline{6-11}
        \cline{12-14}
        & & EPM & ASI & HLD & OCR & ACR & ATR & STU & FPD & OJR & REC & SSR & CRR \\
        \hline

        & Semantic-only & 48.82 & 60.81 & 31.72 & 62.42 & 47.71 & 72.41 & 52.25 & 65.35 & 54.35 & 20.63 & 55.64 & 41.25 \\
        & Recent-only & 52.19 & 58.78 & 40.32 & 91.95 & 67.89 & 81.90 & 67.42 & 75.25 & 78.80 & 18.48 & 56.44 & 43.75 \\
        \rowcolor{gray!15}
        \cellcolor{white}\multirow{-3}{*}{Qwen2.5-VL-7B} & \textbf{Oracle} & \textbf{63.97} & \textbf{66.89} & \textbf{46.77} & \textbf{93.96} & \textbf{72.48} & \textbf{87.93} & \textbf{71.35} & \textbf{81.19} & \textbf{83.70} & \textbf{24.79} & \textbf{65.02} & \textbf{47.08} \\
        \hline

        & Semantic-only & 54.88 & 58.78 & 31.72 & 67.11 & 58.72 & 73.28 & 55.06 & 64.36 & 58.70 & 26.22 & 63.12 & 42.50 \\
        & Recent-only & 52.19 & 55.41 & 40.86 & 94.63 & 80.73 & 81.90 & 67.98 & 75.25 & 83.70 & 22.78 & 64.55 & 45.00 \\
        \rowcolor{gray!15}
        \cellcolor{white}\multirow{-3}{*}{Qwen3-VL-8B} & \textbf{Oracle} & \textbf{67.34} & \textbf{67.57} & \textbf{46.77} & \textbf{95.97} & \textbf{84.40} & \textbf{87.93} & \textbf{70.78} & \textbf{81.19} & \textbf{86.41} & \textbf{30.66} & \textbf{75.67} & \textbf{50.42} \\
        \hline
    \end{tabular}
    \end{adjustbox}
    }
\end{table*}

\section{Additional Analysis of Empirical Observations}
\label{app:empirical_analysis}

We provide additional results for the two empirical observations in Sec.~\ref{sec:empirical_observations}.
We first decompose VTAT to examine where the latency of the encode-first pipeline is spent.
We then report the full 12-task breakdown for Semantic-only, Recent-only, and the per-query Oracle.

\subsection{VTAT Breakdown}
\label{app:vtat_breakdown}

Fig.~\ref{fig:observation1} shows that Qwen2.5-VL-7B requires an average VTAT of 12.18 s under the encode-first pipeline, compared with only 0.243 s for inference TTFT.
Tab.~\ref{tab:vtat_breakdown} further decomposes this latency.

Target-VLM visual encoding takes 7.596 s on average, accounting for 62.36\% of the total VTAT.
Inference TTFT accounts for only 0.243 s (1.99\%).
The remaining 4.342 s (35.65\%) is the residual pipeline overhead after subtracting visual encoding and inference TTFT from VTAT.
Target-VLM visual encoding is therefore the largest component of VTAT under our profiling setup.
This result motivates identifying query evidence before invoking the target model's visual encoder, enabling visual processing to focus only on selected observations.

\subsection{Task-Level Analysis of Fixed Evidence Policies}
\label{app:fixed_policy_tasks}

Tab.~\ref{tab:observation2} reports the aggregate comparison between Semantic-only, Recent-only, and the per-query Oracle.
Tab.~\ref{tab:oracle_tasks} gives the corresponding results for all 12 OVO-Bench tasks.

The relative performance of Semantic-only and Recent-only varies across tasks.
For Qwen3-VL-8B, Recent-only performs better on all six Real-Time tasks, while Semantic-only performs better on EPM, ASI, and REC.
Qwen2.5-VL-7B shows a similar pattern. 
Recent-only performs better on all six Real-Time tasks, whereas Semantic-only is stronger on ASI and REC.

The per-query Oracle improves over both fixed policies on every task for both backbones.
Thus, the aggregate Oracle gains in Tab.~\ref{tab:observation2} are not caused only by differences across task groups.
Even within the same task, different queries can favor different evidence policies.
This supports adapting the semantic-recency trade-off at the query level rather than fixing it globally or by task.

\section{Additional Online Evaluation}
\label{app:additional_online}

\begin{table*}[t]
    \centering
    \caption{
        \textbf{Per-task comparison (\%) on OVO-Bench under online evaluation setting.}
        We report results across the Backward, Real-Time, and Forward task groups.
        HERMES results are taken from its original 6K-token setting.
        ``--'' indicates results not reported.
        Gray rows denote WRWS.
    }
    \label{tab:online_main_tasks}
    \vspace{-2mm}

    \footnotesize
    \renewcommand{\arraystretch}{1.25}
    \setlength{\tabcolsep}{2.8pt}
    \setlength{\extrarowheight}{0.5pt}

    \begin{adjustbox}{max width=\textwidth}
    \begin{tabular}{c|l|ccc|cccccc|ccc}
        \hline
        \multirow{2}{*}{Model} & \multirow{2}{*}{Method} & \multicolumn{3}{c|}{Backward} & \multicolumn{6}{c|}{Real-Time} & \multicolumn{3}{c}{Forward} \\
        \cline{3-5}
        \cline{6-11}
        \cline{12-14}
        & & EPM & ASI & HLD & OCR & ACR & ATR & STU & FPD & OJR & REC & SSR & CRR \\
        \hline

        \multicolumn{14}{c}{\textit{Qwen2.5-VL}} \\
        \hline
        & SimpleStream & \textbf{52.19} & 58.78 & \textbf{40.32} & 91.95 & \textbf{67.89} & \textbf{81.90} & \textbf{67.42} & \textbf{75.25} & \textbf{78.80} & \textbf{18.48} & 56.44 & 43.75 \\
        \rowcolor{gray!12}
        \cellcolor{white}\multirow{-2}{*}{7B} & WRWS & 51.85 & \textbf{60.14} & 39.78 & \textbf{92.62} & \textbf{67.89} & \textbf{81.90} & 66.85 & \textbf{75.25} & 78.26 & 18.34 & \textbf{56.92} & \textbf{45.42} \\
        \hline
        & SimpleStream & 52.19 & \textbf{54.05} & \textbf{40.86} & \textbf{93.96} & 77.06 & \textbf{86.21} & \textbf{71.35} & \textbf{72.28} & \textbf{82.61} & \textbf{19.91} & 62.48 & 43.75 \\
        \rowcolor{gray!12}
        \cellcolor{white}\multirow{-2}{*}{32B} & WRWS & \textbf{54.55} & \textbf{54.05} & 39.25 & 93.29 & \textbf{77.98} & 85.34 & \textbf{71.35} & 71.29 & 81.52 & 19.63 & \textbf{62.96} & \textbf{44.17} \\
        \hline

        \multicolumn{14}{c}{\textit{Qwen3-VL}} \\
        \hline
        & SimpleStream & 51.85 & \textbf{55.41} & 39.25 & 93.29 & \textbf{83.49} & \textbf{83.62} & 67.42 & \textbf{78.22} & 81.52 & 19.34 & 64.55 & 43.75 \\
        \rowcolor{gray!12}
        \cellcolor{white}\multirow{-2}{*}{8B} & WRWS & \textbf{54.55} & 54.05 & \textbf{41.40} & \textbf{94.63} & 80.73 & 81.03 & \textbf{68.54} & 75.25 & \textbf{84.24} & \textbf{22.78} & \textbf{64.86} & \textbf{47.50} \\
        \hline
        & SimpleStream & 59.93 & \textbf{64.19} & \textbf{60.22} & \textbf{95.97} & 88.07 & \textbf{87.93} & \textbf{70.22} & 83.17 & \textbf{87.50} & 20.34 & \textbf{69.95} & 42.08 \\
        \rowcolor{gray!12}
        \cellcolor{white}\multirow{-2}{*}{30B-A3B} & WRWS & \textbf{62.63} & \textbf{64.19} & \textbf{60.22} & 94.63 & \textbf{88.99} & 85.34 & 69.10 & \textbf{85.15} & 86.41 & \textbf{22.78} & \textbf{69.95} & \textbf{44.58} \\
        \hline

        \multicolumn{14}{c}{\textit{Qwen3.5}} \\
        \hline
        & Base & \textbf{63.97} & \textbf{68.24} & 37.63 & 80.54 & 62.39 & 73.28 & 52.81 & 74.26 & 64.67 & \textbf{36.96} & 60.10 & \textbf{60.42} \\
        \rowcolor{gray!12}
        \cellcolor{white}\multirow{-2}{*}{4B} & WRWS & 58.59 & 60.81 & \textbf{51.61} & \textbf{95.97} & \textbf{77.06} & \textbf{81.03} & \textbf{65.17} & \textbf{78.22} & \textbf{81.52} & 20.63 & \textbf{68.84} & 46.25 \\
        \hline
        & Base & \textbf{68.69} & \textbf{79.73} & 39.25 & 84.56 & 71.56 & 76.72 & 58.99 & 79.21 & 66.30 & \textbf{41.12} & 72.34 & \textbf{61.67} \\
        \rowcolor{gray!12}
        \cellcolor{white}\multirow{-2}{*}{35B-A3B} & WRWS & 64.98 & 71.62 & \textbf{58.60} & \textbf{95.97} & \textbf{84.40} & \textbf{82.76} & \textbf{70.79} & \textbf{80.20} & \textbf{82.61} & 20.92 & \textbf{73.29} & 45.00 \\
        \hline

        \multicolumn{14}{c}{\textit{LLaVA-OneVision}} \\
        \hline
        & HERMES & \textbf{61.28} & 58.78 & 26.34 & 72.48 & \textbf{62.39} & 69.83 & 47.75 & \textbf{73.27} & 64.67 & -- & -- & -- \\
        \rowcolor{gray!12}
        \cellcolor{white}\multirow{-2}{*}{7B} & WRWS & 59.93 & \textbf{62.84} & \textbf{29.57} & \textbf{78.52} & 59.63 & \textbf{74.14} & \textbf{52.81} & \textbf{73.27} & \textbf{66.30} & 27.36 & 67.57 & 58.33 \\
        \hline
        & HERMES & \textbf{47.81} & \textbf{47.30} & 9.14 & 57.05 & 49.54 & \textbf{55.17} & 32.58 & \textbf{60.40} & 47.28 & -- & -- & -- \\
        \rowcolor{gray!12}
        \cellcolor{white}\multirow{-2}{*}{0.5B} & WRWS & 47.47 & 44.59 & \textbf{23.66} & \textbf{61.07} & \textbf{55.96} & \textbf{55.17} & \textbf{39.89} & 54.46 & \textbf{52.17} & 15.04 & 62.80 & 45.42 \\
        \hline
    \end{tabular}
    \end{adjustbox}
    \vspace{-5mm}
\end{table*}

\subsection{Detailed Results on the Main Online Benchmarks}
\label{app:online_main_tasks}

Tab.~\ref{tab:online_main_tasks} reports the per-task OVO-Bench results under the main online evaluation setting.
For Qwen2.5-VL-7B, WRWS improves ASI, OCR, SSR, and CRR over SimpleStream, while giving similar results on several other tasks.
At 32B, the improvements appear on EPM, ACR, SSR, and CRR, with comparable performance on ASI and STU.
For Qwen3-VL-8B, WRWS performs better on EPM, HLD, OCR, STU, OJR, REC, SSR, and CRR.
At 30B-A3B, it matches or outperforms SimpleStream on all three Backward tasks and improves ACR, FPD, REC, and CRR.

For Qwen3.5, the gains are concentrated on the Real-Time group.
WRWS improves all six Real-Time tasks at both 4B and 35B-A3B.
It also improves HLD and SSR at both scales, while several other Backward and Forward tasks favor the base model.
This task-level variation is consistent with the different temporal evidence requirements analyzed in Sec.~\ref{sec:adaptive_behavior}.

For LLaVA-OneVision, we compare with the HERMES results reported under its 6K-token setting.
Among the nine Backward and Real-Time tasks reported by HERMES, WRWS obtains higher accuracy on six tasks at 7B and five tasks at 0.5B.

\begin{table*}[t]
    \centering
    \caption{
        \textbf{Comparison with HERMES under the 2-fps online evaluation setting.}
        HERMES uses a 6K visual-token budget.
        WRWS selects $K=6$ observations under its default Qwen3-VL streaming configuration.
        ``B+R'' denotes the average over the Backward and Real-Time groups.
        ``--'' indicates results not reported by HERMES.
        Gray rows denote WRWS.
    }
    \label{tab:hermes_2fps_agg}
    \vspace{-2mm}

    {
    \footnotesize
    \renewcommand{\arraystretch}{1.35}
    \setlength{\extrarowheight}{0.5pt}

    \resizebox{\textwidth}{!}{
    \begin{tabular}{c|l|c|ccccc}
        \hline
        Model & Method & Streaming & Backward & Real-Time & Forward & B+R & All \\
        \hline

        & HERMES & \textbf{81.32} & 46.78 & 73.21 & -- & 60.00 & -- \\
        \rowcolor{gray!12}
        \cellcolor{white}\multirow{-2}{*}{Qwen3-VL-8B}
        & WRWS & 80.39 & \textbf{50.76} & \textbf{80.60} & 44.17 & \textbf{65.68} & 58.51 \\
        \hline

        & HERMES & 78.40 & \textbf{54.00} & 71.90 & -- & 62.95 & -- \\
        \rowcolor{gray!12}
        \cellcolor{white}\multirow{-2}{*}{Qwen3-VL-4B}
        & WRWS & \textbf{80.07} & 52.24 & \textbf{78.64} & 41.26 & \textbf{65.44} & 57.38 \\
        \hline
    \end{tabular}
    }
    }
\end{table*}

\begin{table*}[t]
    \centering
    \caption{
        \textbf{Per-task comparison with HERMES under the 2-fps online evaluation setting.}
        HERMES uses a 6K visual-token budget.
        ``--'' indicates results not reported.
        Gray rows denote WRWS.
    }
    \label{tab:hermes_2fps_tasks}
    \vspace{-2mm}

    {
    \footnotesize
    \renewcommand{\arraystretch}{1.25}
    \setlength{\tabcolsep}{2.8pt}
    \setlength{\extrarowheight}{0.5pt}

    \begin{adjustbox}{max width=\textwidth}
    \begin{tabular}{c|l|ccc|cccccc|ccc}
        \hline
        \multirow{2}{*}{Model} & \multirow{2}{*}{Method} & \multicolumn{3}{c|}{Backward} & \multicolumn{6}{c|}{Real-Time} & \multicolumn{3}{c}{Forward} \\
        \cline{3-5}
        \cline{6-11}
        \cline{12-14}
        & & EPM & ASI & HLD & OCR & ACR & ATR & STU & FPD & OJR & REC & SSR & CRR \\
        \hline

        & HERMES & \textbf{54.88} & \textbf{70.95} & 14.52 & 85.91 & 71.56 & \textbf{81.03} & 56.74 & 72.28 & 71.74 & -- & -- & -- \\
        \rowcolor{gray!12}
        \cellcolor{white}\multirow{-2}{*}{Qwen3-VL-8B} & WRWS & 53.87 & 55.41 & \textbf{43.01} & \textbf{94.63} & \textbf{83.49} & 79.31 & \textbf{71.35} & \textbf{73.27} & \textbf{81.52} & 20.06 & 66.61 & 45.83 \\
        \hline

        & HERMES & 54.55 & \textbf{62.84} & 44.62 & 81.88 & 73.39 & \textbf{77.59} & 58.43 & 73.27 & 66.85 & -- & -- & -- \\
        \rowcolor{gray!12}
        \cellcolor{white}\multirow{-2}{*}{Qwen3-VL-4B} & WRWS & \textbf{55.22} & 54.73 & \textbf{46.77} & \textbf{93.96} & \textbf{77.98} & 76.72 & \textbf{67.42} & \textbf{74.26} & \textbf{81.52} & 17.19 & 65.34 & 41.25 \\
        \hline
    \end{tabular}
    \end{adjustbox}
    }
\end{table*}

\subsection{Comparison with HERMES under the 2-fps Setting}
\label{app:hermes_2fps}

We further compare WRWS with HERMES under the 2-fps online evaluation setting used in HERMES.
As shown in Tab.~\ref{tab:hermes_2fps_agg}, WRWS improves the Real-Time average from 73.21 to 80.60 on Qwen3-VL-8B and from 71.90 to 78.64 on Qwen3-VL-4B.
The corresponding B+R averages increase from 60.00 to 65.68 and from 62.95 to 65.44.
On StreamingBench, WRWS is slightly lower than HERMES at 8B (80.39 vs.\ 81.32) and higher at 4B (80.07 vs.\ 78.40).

Tab.~\ref{tab:hermes_2fps_tasks} provides the per-task OVO-Bench results.
On Qwen3-VL-8B, WRWS improves six of the nine Backward and Real-Time tasks, with particularly large gains on HLD, OCR, ACR, STU, and OJR.
On Qwen3-VL-4B, WRWS improves seven of the nine reported tasks, including EPM, HLD, OCR, ACR, STU, FPD, and OJR.
HERMES remains stronger on ASI and ATR at both model scales.

\begin{table*}[t]
    \centering
    \caption{
        \textbf{Performance under different target-model visual-processing budgets.}
        We vary the number of selected observations $K$ for Qwen3-VL-30B-A3B.
        ``B+R'' denotes the average over the Backward and Real-Time groups.
        The gray row denotes the default setting ($K=6$).
    }
    \label{tab:budget_agg}
    \vspace{-2mm}

    {
    \footnotesize
    \renewcommand{\arraystretch}{1.15}
    \setlength{\extrarowheight}{0.5pt}

    \resizebox{\textwidth}{!}{
    \begin{tabular}{c|l|c|c|ccccc}
        \hline
        Model & Method & $K$ & Streaming & Backward & Real-Time & Forward & B+R & All \\
        \hline

        & WRWS & 2  & 78.83 & 60.46 & 83.03 & 42.91 & 71.74 & 62.13 \\
        & WRWS & 4  & 83.51 & 62.17 & \textbf{85.33} & 45.26 & \textbf{73.75} & 64.25 \\

        \rowcolor{gray!12}
        \cellcolor{white}
        & WRWS & 6  & 83.71 & 62.35 & 84.94 & 45.77 & 73.59 & 64.35 \\

        & WRWS & 8  & 83.67 & 62.35 & 84.62 & 47.33 & 73.49 & 64.77 \\

        \multirow{-5}{*}{30B-A3B}
        & WRWS & 16 & \textbf{84.31} & \textbf{64.53} & 81.29 & \textbf{50.41} & 72.91 & \textbf{65.41} \\
        \hline
    \end{tabular}
    }
    }
\end{table*}

\begin{table*}[t]
    \centering
    \caption{
        \textbf{Per-task performance under different target-model visual-processing budgets.}
        We vary the number of selected observations $K$ for Qwen3-VL-30B-A3B.
        The gray row denotes the default setting ($K=6$).
    }
    \label{tab:budget_tasks}
    \vspace{-2mm}

    {
    \footnotesize
    \renewcommand{\arraystretch}{1.25}
    \setlength{\tabcolsep}{2.8pt}
    \setlength{\extrarowheight}{0.5pt}

    \begin{adjustbox}{max width=\textwidth}
    \begin{tabular}{c|l|c|ccc|cccccc|ccc}
        \hline
        \multirow{2}{*}{Model} & \multirow{2}{*}{Method} & \multirow{2}{*}{$K$} & \multicolumn{3}{c|}{Backward} & \multicolumn{6}{c|}{Real-Time} & \multicolumn{3}{c}{Forward} \\
        \cline{4-6}
        \cline{7-12}
        \cline{13-15}
        & & & EPM & ASI & HLD & OCR & ACR & ATR & STU & FPD & OJR & REC & SSR & CRR \\
        \hline

        & WRWS & 2 & 57.24 & 65.54 & 58.60 & 94.63 & 79.82 & 84.48 & 69.66 & 83.17 & 86.41 & 18.05 & 69.00 & 41.67 \\
        & WRWS & 4 & 60.61 & 66.22 & 59.68 & \textbf{95.97} & 88.07 & \textbf{87.07} & \textbf{70.22} & 83.17 & \textbf{87.50} & 20.34 & \textbf{70.43} & 45.00 \\
        \rowcolor{gray!12}
        \cellcolor{white} & WRWS & 6 & 62.63 & 64.19 & \textbf{60.22} & 94.63 & \textbf{88.99} & 85.34 & 69.10 & \textbf{85.15} & 86.41 & 22.78 & 69.95 & 44.58 \\
        & WRWS & 8 & 61.95 & 67.57 & 57.53 & 95.30 & 87.16 & 84.48 & \textbf{70.22} & 84.16 & 86.41 & 24.07 & \textbf{70.43} & 47.50 \\
        \multirow{-5}{*}{30B-A3B} & WRWS & 16 & \textbf{66.33} & \textbf{70.27} & 56.99 & 93.96 & 77.98 & 81.03 & 67.98 & 84.16 & 82.61 & \textbf{27.36} & 70.11 & \textbf{53.75} \\
        \hline
    \end{tabular}
    \end{adjustbox}
    }
\end{table*}

\subsection{Effect of the Target-Model Visual-Processing Budget}
\label{app:token_budget}

We examine the effect of the target-model visual-processing budget by varying $K$ from 2 to 16 on Qwen3-VL-30B-A3B.
As shown in Tab.~\ref{tab:budget_agg}, increasing $K$ from 2 to 4 improves the OVO-Bench average from 62.13 to 64.25.
The overall score then changes only modestly as $K$ increases from 4 to 8, reaching 64.35 at the default $K=6$ and 64.77 at $K=8$.
With $K=16$, the overall score further increases to 65.41.
The gains mainly come from the Backward and Forward groups, whose averages increase to 64.53 and 50.41.
In contrast, the Real-Time average decreases from 84.94 at $K=6$ to 81.29 at $K=16$.
WRWS therefore remains effective across a range of visual-processing budgets, while the preferred budget varies with the temporal evidence requirements of different tasks.

The per-task results in Tab.~\ref{tab:budget_tasks} show the same trend.
EPM, ASI, REC, and CRR generally benefit from a larger $K$, while several Real-Time tasks perform best with smaller or intermediate budgets.
For example, ATR reaches its highest score at $K=4$, while ACR and FPD peak at $K=6$.
Overall, the results remain close for $K=4$, $6$, and $8$, indicating that the default setting is not sensitive to a narrow choice of the visual-processing budget.

\begin{table*}[t]
    \centering
    \caption{
        \textbf{Comparison with SimpleStream across model scales.}
        For each target model, we compare against the best reported SimpleStream configuration, which may use a model-specific evidence budget $K$.
        WRWS keeps $K=4$ for all Qwen2.5-VL models and $K=6$ for all Qwen3-VL models.
        ``B+R'' denotes the average over the Backward and Real-Time groups.
        Gray rows denote WRWS.
    }
    \label{tab:scaling_simplestream_agg}
    \vspace{-2mm}

    {
    \footnotesize
    \renewcommand{\arraystretch}{1.25}
    \setlength{\extrarowheight}{0.5pt}
    \setlength{\tabcolsep}{5pt}

    \begin{tabularx}{\textwidth}{c|l|c|C|CCCCC}
        \hline
        Model & Method & $K$ & Streaming & Backward & Real-Time & Forward & B+R & All \\
        \hline

        \multicolumn{9}{c}{\textit{Qwen2.5-VL}} \\
        \hline
        & SimpleStream & 4 & 77.83 & \textbf{48.50} & \textbf{74.64} & 43.19 & \textbf{61.57} & 55.44 \\
        \rowcolor{gray!12}
        \cellcolor{white}\multirow{-2}{*}{3B} & WRWS & 4 & \textbf{78.07} & 47.27 & 74.56 & \textbf{45.21} & 60.91 & \textbf{55.68} \\
        \hline

        & SimpleStream & 4 & \textbf{80.31} & 49.03 & \textbf{80.58} & 42.05 & \textbf{64.81} & \textbf{57.22} \\
        \rowcolor{gray!12}
        \cellcolor{white}\multirow{-2}{*}{32B} & WRWS & 4 & 80.23 & \textbf{49.28} & 80.13 & \textbf{42.25} & 64.65 & 57.22 \\
        \hline

        & SimpleStream & 16 & \textbf{84.31} & \textbf{61.46} & 79.52 & \textbf{42.07} & 70.49 & \textbf{61.02} \\
        \rowcolor{gray!12}
        \cellcolor{white}\multirow{-2}{*}{72B} & WRWS & 4 & 82.35 & 60.89 & \textbf{81.25} & 40.72 & \textbf{71.07} & 60.95 \\
        \hline

        \multicolumn{9}{c}{\textit{Qwen3-VL}} \\
        \hline
        & SimpleStream & 4 & 76.55 & \textbf{44.66} & \textbf{75.87} & 43.77 & \textbf{60.27} & \textbf{54.77} \\
        \rowcolor{gray!12}
        \cellcolor{white}\multirow{-2}{*}{2B} & WRWS & 6 & \textbf{76.91} & 44.60 & 74.51 & \textbf{44.64} & 59.55 & 54.58 \\
        \hline

        & SimpleStream & 16 & \textbf{80.79} & \textbf{54.72} & 77.20 & \textbf{42.99} & \textbf{65.96} & \textbf{58.31} \\
        \rowcolor{gray!12}
        \cellcolor{white}\multirow{-2}{*}{4B} & WRWS & 6 & 80.51 & 50.81 & \textbf{78.51} & 42.44 & 64.66 & 57.25 \\
        \hline

        & SimpleStream & 8 & \textbf{83.31} & 64.88 & 83.05 & \textbf{45.64} & 73.96 & \textbf{64.52} \\
        \rowcolor{gray!12}
        \cellcolor{white}\multirow{-2}{*}{32B} & WRWS & 6 & 83.03 & \textbf{65.15} & \textbf{83.11} & 44.27 & \textbf{74.13} & 64.18 \\
        \hline

        & SimpleStream & 4 & 83.51 & 61.45 & \textbf{85.48} & 44.12 & 73.46 & 63.68 \\
        \rowcolor{gray!12}
        \cellcolor{white}\multirow{-2}{*}{30B-A3B} & WRWS & 6 & \textbf{83.71} & \textbf{62.35} & 84.94 & \textbf{45.77} & \textbf{73.59} & \textbf{64.35} \\
        \hline
    \end{tabularx}
    }
\end{table*}

\begin{table*}[t]
    \centering
    \caption{
        \textbf{Per-task comparison with SimpleStream across model scales.}
        SimpleStream uses its best reported model-specific evidence budget $K$, while WRWS keeps $K=4$ for Qwen2.5-VL and $K=6$ for Qwen3-VL.
        Gray rows denote WRWS.
    }
    \label{tab:scaling_simplestream_tasks}
    \vspace{-2mm}
    {
    \footnotesize
    \renewcommand{\arraystretch}{1.25}
    \setlength{\tabcolsep}{2.8pt}
    \setlength{\extrarowheight}{0.5pt}

    \begin{adjustbox}{max width=\textwidth}
    \begin{tabular}{c|l|c|ccc|cccccc|ccc}
        \hline
        \multirow{2}{*}{Model} & \multirow{2}{*}{Method} & \multirow{2}{*}{$K$} & \multicolumn{3}{c|}{Backward} & \multicolumn{6}{c|}{Real-Time} & \multicolumn{3}{c}{Forward} \\
        \cline{4-6}
        \cline{7-12}
        \cline{13-15}
        & & & EPM & ASI & HLD & OCR & ACR & ATR & STU & FPD & OJR & REC & SSR & CRR \\
        \hline

        \multicolumn{15}{c}{\textit{Qwen2.5-VL}} \\
        \hline
        & SimpleStream & 4 & \textbf{52.19} & \textbf{56.76} & \textbf{36.56} & 88.59 & 73.39 & \textbf{75.86} & \textbf{59.55} & \textbf{73.27} & \textbf{77.17} & 17.62 & 68.20 & 43.75 \\
        \rowcolor{gray!12}
        \cellcolor{white}\multirow{-2}{*}{3B} & WRWS & 4 & 51.85 & 56.08 & 33.87 & \textbf{89.26} & \textbf{74.31} & \textbf{75.86} & \textbf{59.55} & 72.28 & 76.09 & \textbf{17.77} & \textbf{68.68} & \textbf{49.17} \\
        \hline

        & SimpleStream & 4 & 52.19 & \textbf{54.05} & \textbf{40.86} & \textbf{93.96} & 77.06 & \textbf{86.21} & \textbf{71.35} & \textbf{72.28} & \textbf{82.61} & \textbf{19.91} & 62.48 & 43.75 \\
        \rowcolor{gray!12}
        \cellcolor{white}\multirow{-2}{*}{32B} & WRWS & 4 & \textbf{54.55} & \textbf{54.05} & 39.25 & 93.29 & \textbf{77.98} & 85.34 & \textbf{71.35} & 71.29 & 81.52 & 19.63 & \textbf{62.96} & \textbf{44.17} \\
        \hline

        & SimpleStream & 16 & \textbf{64.65} & \textbf{72.97} & 46.77 & 93.96 & 70.64 & 82.76 & 67.42 & \textbf{79.21} & \textbf{83.15} & \textbf{23.07} & 55.64 & \textbf{47.50} \\
        \rowcolor{gray!12}
        \cellcolor{white}\multirow{-2}{*}{72B} & WRWS & 4 & 59.60 & 68.24 & \textbf{54.84} & \textbf{95.30} & \textbf{83.49} & \textbf{83.62} & \textbf{69.66} & 72.28 & \textbf{83.15} & 18.05 & \textbf{59.94} & 44.17 \\
        \hline

        \multicolumn{15}{c}{\textit{Qwen3-VL}} \\
        \hline
        & SimpleStream & 4 & 51.85 & \textbf{52.03} & \textbf{30.11} & 92.62 & \textbf{76.15} & \textbf{80.17} & \textbf{51.12} & 75.25 & \textbf{79.89} & 18.62 & 71.86 & 40.83 \\
        \rowcolor{gray!12}
        \cellcolor{white}\multirow{-2}{*}{2B} & WRWS & 6 & \textbf{52.19} & \textbf{52.03} & 29.57 & \textbf{93.29} & 70.64 & 77.59 & \textbf{51.12} & \textbf{77.23} & 77.17 & \textbf{20.06} & \textbf{72.18} & \textbf{41.67} \\
        \hline

        & SimpleStream & 16 & \textbf{60.61} & \textbf{62.16} & 41.40 & 92.62 & 75.23 & \textbf{77.59} & 62.92 & \textbf{78.22} & 76.63 & \textbf{19.77} & 62.96 & \textbf{46.25} \\
        \rowcolor{gray!12}
        \cellcolor{white}\multirow{-2}{*}{4B} & WRWS & 6 & 55.22 & 54.73 & \textbf{42.47} & \textbf{93.29} & \textbf{77.06} & 75.00 & \textbf{67.42} & 76.24 & \textbf{82.07} & 18.34 & \textbf{67.73} & 41.25 \\
        \hline

        & SimpleStream & 8 & 55.56 & \textbf{67.57} & \textbf{71.51} & \textbf{95.30} & \textbf{85.32} & \textbf{85.34} & 72.47 & \textbf{77.23} & 82.61 & \textbf{18.05} & \textbf{71.38} & \textbf{47.50} \\
        \rowcolor{gray!12}
        \cellcolor{white}\multirow{-2}{*}{32B} & WRWS & 6 & \textbf{56.90} & \textbf{67.57} & 70.97 & \textbf{95.30} & 84.40 & \textbf{85.34} & \textbf{73.60} & 75.25 & \textbf{84.78} & 15.33 & 71.22 & 46.25 \\
        \hline

        & SimpleStream & 4 & 59.93 & \textbf{64.19} & \textbf{60.22} & \textbf{95.97} & 88.07 & \textbf{87.93} & \textbf{70.22} & 83.17 & \textbf{87.50} & 20.34 & \textbf{69.95} & 42.08 \\
        \rowcolor{gray!12}
        \cellcolor{white}\multirow{-2}{*}{30B-A3B} & WRWS & 6 & \textbf{62.63} & \textbf{64.19} & \textbf{60.22} & 94.63 & \textbf{88.99} & 85.34 & 69.10 & \textbf{85.15} & 86.41 & \textbf{22.78} & \textbf{69.95} & \textbf{44.58} \\
        \hline
    \end{tabular}
    \end{adjustbox}
    }
\end{table*}

\subsection{Results across Model Scales}
\label{app:scale}

We further compare WRWS with SimpleStream across additional model scales.
For WRWS, we keep the evidence budget fixed within each model family, using $K=4$ for Qwen2.5-VL and $K=6$ for Qwen3-VL.
\textbf{SimpleStream uses its best reported model-specific configuration.}

As shown in Tab.~\ref{tab:scaling_simplestream_agg}, WRWS remains close to SimpleStream across all evaluated scales.
On Qwen2.5-VL, WRWS improves the overall OVO-Bench score at 3B, matches SimpleStream at 32B, and differs by only 0.07 points at 72B.
The 72B comparison is particularly notable because WRWS uses $K=4$, compared with $K=16$ for SimpleStream.
For Qwen3-VL, WRWS remains close to SimpleStream across the four evaluated scales, with differences ranging from $-1.06$ to $+0.67$ points.
At 30B-A3B, WRWS exceeds SimpleStream by 0.67 points.

The per-task results in Tab.~\ref{tab:scaling_simplestream_tasks} further show that the differences between WRWS and SimpleStream vary across tasks and model sizes.
WRWS improves several Forward tasks, while SimpleStream remains stronger on some Backward and Real-Time tasks.
Overall, WRWS maintains comparable performance across model scales without model-specific adjustment of $K$ within each model family.

\begin{table*}[t]
    \centering
    \caption{
        \textbf{Comparison of adaptive and fixed evidence allocation on OVO-Bench.}
        Fixed $\alpha$ values interpolate between Semantic-only ($\alpha=0$)
        and Recent-only ($\alpha=1$).
        Adaptive denotes the query-adaptive allocation used by WRWS.
        ``B+R'' denotes the average over the Backward and Real-Time groups.
        Gray rows denote Adaptive.
    }
    \label{tab:adaptive_allocation_agg}
    \vspace{-2mm}
    {
    \renewcommand{\arraystretch}{1.05}
    \setlength{\extrarowheight}{0.5pt}

    \resizebox{\textwidth}{!}{
    \begin{tabular}{c|l|ccccc}
        \hline
        Model & Allocation & Backward & Real-Time & Forward & B+R & All \\
        \hline

        & $\alpha=0.00$ (Semantic-only)
        & 47.12 & 59.08 & 39.17 & 53.10 & 48.46 \\

        & $\alpha=0.25$
        & 47.97 & 64.46 & \textbf{40.72} & 56.21 & 51.05 \\

        & $\alpha=0.50$
        & 50.22 & 68.34 & 39.96 & 59.28 & 52.84 \\

        & $\alpha=0.75$
        & 49.97 & 73.95 & 38.58 & 61.96 & 54.16 \\

        & $\alpha=1.00$ (Recent-only)
        & 50.43 & \textbf{77.20} & 39.56 & 63.82 & 55.73 \\

        \rowcolor{gray!12}
        \cellcolor{white}\multirow{-6}{*}{Qwen2.5-VL-7B}
        & Adaptive
        & \textbf{50.59} & 77.13 & 40.23
        & \textbf{63.86} & \textbf{55.98} \\
        \hline

        & $\alpha=0.00$ (Semantic-only)
        & 48.46 & 62.87 & 43.95 & 55.67 & 51.76 \\

        & $\alpha=0.25$
        & 48.80 & 69.72 & 45.90 & 59.26 & 54.81 \\

        & $\alpha=0.50$
        & 49.56 & 73.32 & 45.57 & 61.44 & 56.15 \\

        & $\alpha=0.75$
        & 49.90 & 78.39 & \textbf{46.47} & 64.14 & 58.25 \\

        & $\alpha=1.00$ (Recent-only)
        & 49.49 & 80.70 & 44.11 & 65.09 & 58.10 \\

        \rowcolor{gray!12}
        \cellcolor{white}\multirow{-6}{*}{Qwen3-VL-8B}
        & Adaptive
        & \textbf{50.00} & \textbf{80.74} & 45.05
        & \textbf{65.37} & \textbf{58.59} \\
        \hline

    \end{tabular}
    }
    }
\end{table*}

\begin{table*}[t]
    \centering
    \caption{
        \textbf{Per-task comparison of adaptive and fixed evidence allocation on OVO-Bench.}
        Fixed $\alpha$ values range from Semantic-only ($\alpha=0$) to Recent-only ($\alpha=1$).
        Gray rows denote Adaptive.
        Best results for each backbone are shown in \textbf{bold}.
    }
    \label{tab:adaptive_allocation_tasks}
    \vspace{-2mm}

    {
    \footnotesize
    \renewcommand{\arraystretch}{1.25}
    \setlength{\tabcolsep}{2.8pt}
    \setlength{\extrarowheight}{0.5pt}

    \begin{adjustbox}{max width=\textwidth}
    \begin{tabular}{c|l|ccc|cccccc|ccc}
        \hline
        \multirow{2}{*}{Model} & \multirow{2}{*}{Allocation} & \multicolumn{3}{c|}{Backward} & \multicolumn{6}{c|}{Real-Time} & \multicolumn{3}{c}{Forward} \\
        \cline{3-5}
        \cline{6-11}
        \cline{12-14}
        & & EPM & ASI & HLD & OCR & ACR & ATR & STU & FPD & OJR & REC & SSR & CRR \\
        \hline

        & $\alpha=0.00$ & 48.82 & 60.81 & 31.72 & 62.42 & 47.71 & 72.41 & 52.25 & 65.35 & 54.35 & 20.63 & 55.64 & 41.25 \\
        & $\alpha=0.25$ & 52.19 & 59.46 & 32.26 & 74.50 & 49.54 & 74.14 & 56.74 & 69.31 & 62.50 & \textbf{20.92} & 57.07 & 44.17 \\
        & $\alpha=0.50$ & \textbf{54.21} & \textbf{64.19} & 32.26 & 81.21 & 58.72 & 75.86 & 59.55 & 67.33 & 67.39 & 20.20 & 56.76 & 42.92 \\
        & $\alpha=0.75$ & 52.53 & 60.81 & 36.56 & 89.93 & 66.06 & 80.17 & 64.04 & 72.28 & 71.20 & 18.19 & \textbf{57.55} & 40.00 \\
        & $\alpha=1.00$ & 52.19 & 58.78 & \textbf{40.32} & 91.95 & \textbf{67.89} & \textbf{81.90} & \textbf{67.42} & \textbf{75.25} & \textbf{78.80} & 18.48 & 56.44 & 43.75 \\
        \rowcolor{gray!12}
        \cellcolor{white}\multirow{-6}{*}{Qwen2.5-VL-7B} & Adaptive & 51.85 & 60.14 & 39.78 & \textbf{92.62} & \textbf{67.89} & \textbf{81.90} & 66.85 & \textbf{75.25} & 78.26 & 18.34 & 56.92 & \textbf{45.42} \\
        \hline

        & $\alpha=0.00$ & 54.88 & 58.78 & 31.72 & 67.11 & 58.72 & 73.28 & 55.06 & 64.36 & 58.70 & \textbf{26.22} & 63.12 & 42.50 \\
        & $\alpha=0.25$ & \textbf{55.89} & 58.78 & 31.72 & 77.18 & 68.81 & 76.72 & 60.11 & 70.30 & 65.22 & \textbf{26.22} & 67.73 & 43.75 \\
        & $\alpha=0.50$ & 55.22 & \textbf{60.14} & 33.33 & 83.89 & 74.31 & 77.59 & 65.17 & 68.32 & 70.65 & 24.93 & \textbf{68.04} & 43.75 \\
        & $\alpha=0.75$ & 54.21 & 59.46 & 36.02 & 91.95 & 79.82 & 79.31 & 66.29 & \textbf{75.25} & 77.72 & 23.50 & 65.50 & \textbf{50.42} \\
        & $\alpha=1.00$ & 52.19 & 55.41 & 40.86 & \textbf{94.63} & \textbf{80.73} & \textbf{81.90} & 67.98 & \textbf{75.25} & 83.70 & 22.78 & 64.55 & 45.00 \\
        \rowcolor{gray!12}
        \cellcolor{white}\multirow{-6}{*}{Qwen3-VL-8B} & Adaptive & 54.55 & 54.05 & \textbf{41.40} & \textbf{94.63} & \textbf{80.73} & 81.03 & \textbf{68.54} & \textbf{75.25} & \textbf{84.24} & 22.78 & 64.86 & 47.50 \\
        \hline
    \end{tabular}
    \end{adjustbox}
    }
\end{table*}

\section{Additional Analysis of Adaptive Evidence Allocation}
\label{app:additional_ablation}

\subsection{Adaptive vs. Fixed Evidence Allocation}

We further compare adaptive allocation with fixed $\alpha$ values on Qwen2.5-VL-7B and Qwen3-VL-8B.
As shown in Tab.~\ref{tab:adaptive_allocation_agg}, the best fixed $\alpha$ differs across backbones.
Qwen2.5-VL-7B performs best with $\alpha=1.0$, reaching 55.73, whereas Qwen3-VL-8B favors $\alpha=0.75$, reaching 58.25.
Adaptive allocation does not require selecting a model-specific $\alpha$ and achieves higher overall scores on both backbones, with 55.98 on Qwen2.5-VL-7B and 58.59 on Qwen3-VL-8B.

Tab.~\ref{tab:adaptive_allocation_tasks} provides the corresponding per-task results.
The preferred fixed $\alpha$ varies across tasks on both backbones.
For Qwen2.5-VL-7B, EPM and ASI favor $\alpha=0.5$ and REC favors $\alpha=0.25$, whereas several Real-Time tasks prefer stronger recency weighting.
Qwen3-VL-8B shows a similar trend, with smaller $\alpha$ values favored by EPM and ASI and stronger recency weighting by several Real-Time tasks.
The task-dependent optima indicate that no single fixed semantic--recency trade-off works uniformly across tasks.

\begin{table}[t]
    \centering
    \caption{
        \textbf{Uncertainty-conditioned relative utility of semantic and recent evidence on OVO-Bench.}
        Queries are partitioned into five retrieval-uncertainty quintiles independently within each task and then micro-aggregated across tasks.
        $\Delta_{\mathrm{S-R}}$ denotes Semantic-only accuracy minus Recent-only accuracy in percentage points.
        Confidence intervals are obtained using 5,000 within-task stratified bootstrap repetitions.
        Bold indicates confidence intervals that exclude zero.
    }
    \label{tab:uncertainty_policy_utility}
    \vspace{-2mm}
    {
    \footnotesize
    \renewcommand{\arraystretch}{1.25}
    \setlength{\tabcolsep}{4.0pt}
    \setlength{\extrarowheight}{0.5pt}

    \begin{adjustbox}{max width=\columnwidth}
    \begin{tabular}{l|c|cc|cc}
        \hline
        Uncertainty & \# Queries & Semantic-only & Recent-only & $\Delta_{\mathrm{S-R}}$ & 95\% CI \\
        \hline
        Q1 (lowest) & 612 & 62.09 & 62.91 & $-0.82$ & [$-3.93$, $2.45$] \\
        Q2 & 609 & 52.71 & 55.83 & $-3.12$ & [$-6.57$, $0.16$] \\
        Q3 & 608 & 49.18 & 52.63 & $-3.45$ & [$-6.91$, $0.00$] \\
        Q4 & 605 & 43.14 & 52.73 & $\mathbf{-9.59}$ & $\mathbf{[-13.39,\,-5.79]}$ \\
        Q5 (highest) & 601 & 41.60 & 49.42 & $\mathbf{-7.82}$ & $\mathbf{[-11.48,\,-4.16]}$ \\
        \hline
    \end{tabular}
    \end{adjustbox}
    }
\end{table}

\subsection{Uncertainty-Conditioned Utility of Semantic and Recent Evidence}
As retrieval uncertainty increases, Recent-only exhibits a larger advantage over Semantic-only.
In the lowest-uncertainty quintile, the two policies are nearly tied (62.09 vs.\ 62.91, $\Delta_{\mathrm{S-R}}=-0.82$ points). The gap widens to $-9.59$ and $-7.82$ points in the two highest-uncertainty quintiles.
This trend suggests that retrieval uncertainty is associated with the relative downstream utility of semantic and recent evidence. Semantic retrieval remains more competitive under low uncertainty, while high uncertainty favors a stronger recency prior.

\begin{table*}[t]

    \centering

    \caption{
        \textbf{Detailed efficiency comparison with SimpleStream on OVO-Bench.}
        VTAT measures the elapsed time from the start of processing the first video observation to the generation of the first answer token, while inference TTFT measures the elapsed time from the start of target Video-LLM inference for answering to the generation of the first answer token.
        ``Ratio'' reports WRWS relative to SimpleStream, where a lower VTAT ratio indicates larger latency savings.
        GPU memory is measured over the full evaluation.
    }

    \label{tab:efficiency_detail}
    \vspace{-2mm}

    {
    \renewcommand{\arraystretch}{1.25}
    \setlength{\extrarowheight}{0.5pt}

    \resizebox{\textwidth}{!}{
    \begin{tabular}{l|c|ccc|ccc|cc}
        \hline
        \multirow{2}{*}{Group}
        & \multirow{2}{*}{\# Samples}
        & \multicolumn{3}{c|}{VTAT (s)}
        & \multicolumn{3}{c|}{Inference TTFT (s)}
        & \multicolumn{2}{c}{GPU Memory (GB)} \\
        \cline{3-5}
        \cline{6-8}
        \cline{9-10}

        &
        & WRWS & SimpleStream & Ratio
        & WRWS & SimpleStream & Ratio
        & WRWS & SimpleStream \\
        \hline

        Overall
        & 3035
        & \textbf{5.84} & 12.18 & \textbf{47.93\%}
        & 0.2431 & 0.2429 & 100.08\%
        & \multirow{4}{*}{20.76}
        & \multirow{4}{*}{\textbf{17.28}} \\

        Backward
        & 631
        & \textbf{6.37} & 15.56 & \textbf{40.92\%}
        & 0.2788 & 0.2793 & 99.82\%
        & & \\

        Real-Time
        & 837
        & \textbf{7.17} & 16.50 & \textbf{43.44\%}
        & 0.2863 & 0.2866 & 99.91\%
        & & \\

        Forward
        & 1567
        & \textbf{4.92} & 8.51 & \textbf{57.74\%}
        & 0.2057 & 0.2050 & 100.35\%
        & & \\

        \hline
    \end{tabular}
    }
    }

\end{table*}

\section{Additional Efficiency Analysis}
\label{app:efficiency}

Tab.~\ref{tab:efficiency_detail} provides a detailed efficiency comparison between WRWS and SimpleStream on OVO-Bench using Qwen2.5-VL-7B.
Across all 3,035 samples, WRWS reduces the average VTAT from 12.18 s to 5.84 s, corresponding to 47.93\% of the SimpleStream latency.
The reduction is consistent across the three task groups, with VTAT ratios of 40.92\%, 43.44\%, and 57.74\% for Backward, Real-Time, and Forward tasks, respectively.
The largest relative reduction is observed on Backward tasks, while Forward tasks retain a larger fraction of the baseline latency.

Inference TTFT measures the elapsed time from the start of target Video-LLM inference for answering to the generation of the first answer token.
The overall TTFT is 0.2431 s for WRWS and 0.2429 s for SimpleStream, with similarly small differences across the three task groups.
After the query arrives, both methods perform the same target-model visual encoding and answer generation under the same evidence budget.
Thus, their nearly identical TTFT is expected, and the remaining sub-millisecond differences are likely attributable to normal runtime measurement noise.

WRWS uses 20.76 GB of GPU memory, compared with 17.28 GB for SimpleStream.
The additional memory mainly comes from the SigLIP retrieval encoder and the cached representations of historical observations.
In our implementation, these representations are kept on GPU throughout evaluation, so the retrieval cache grows with the observed video history.
The per-observation storage remains lightweight, as WRWS retains only a single global retrieval embedding for each observation, whereas target-model visual representations typically contain multiple visual tokens per observation.
Furthermore, the retrieval cache can also be \textbf{offloaded to CPU memory}, as commonly done for historical representations in streaming systems, to further reduce its GPU footprint.
We report the all-GPU setting because it avoids additional host--device transfers and offloading logic in the serving pipeline.
Like SimpleStream, WRWS forwards only the selected $K$ observations to the target Video-LLM for each query, regardless of the observed video length.
Its target-model visual context therefore remains bounded by the evidence budget.
When the retrieval cache is offloaded to CPU memory, WRWS also maintains a length-stable GPU-memory profile similar to SimpleStream, in contrast to methods whose GPU footprint grows with the accumulated visual history, such as TimeChat-Online~\citep{timechatonline}.
Under the all-GPU setting reported here, WRWS reduces average VTAT by 52.1\% at the cost of an additional 3.48\,GB of GPU memory.

\section{Qualitative Analysis and Failure Cases}
\label{app:qualitative}

\begin{figure*}[t]
    \centering

    \begin{subfigure}[t]{0.49\textwidth}
        \centering
        \includegraphics[width=\linewidth]{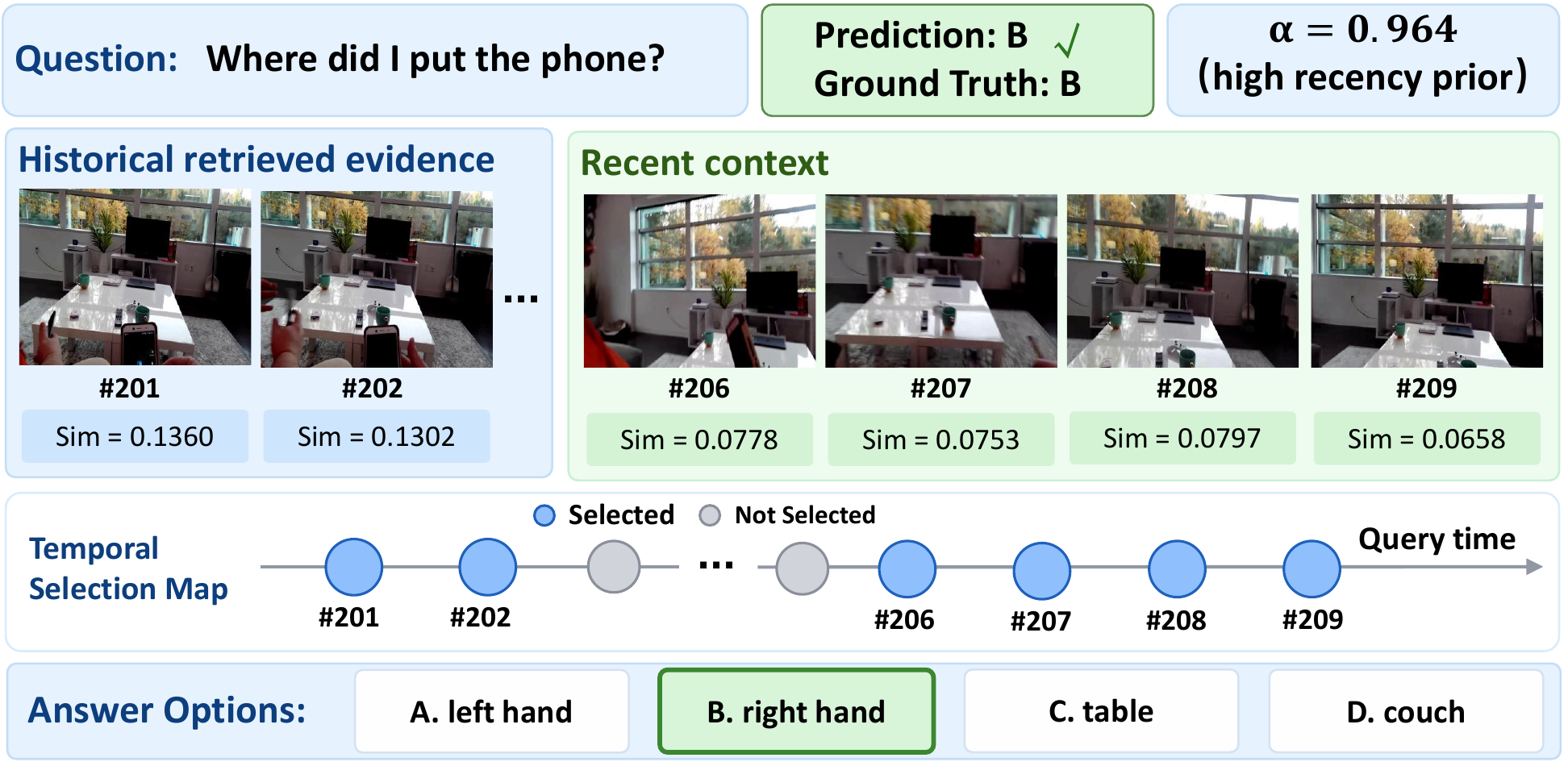}
        \caption{
            \textbf{Retrieving history while keeping recent context.}
        }
        \label{fig:caseA}
    \end{subfigure}
    \hfill
    \begin{subfigure}[t]{0.49\textwidth}
        \centering
        \includegraphics[width=\linewidth]{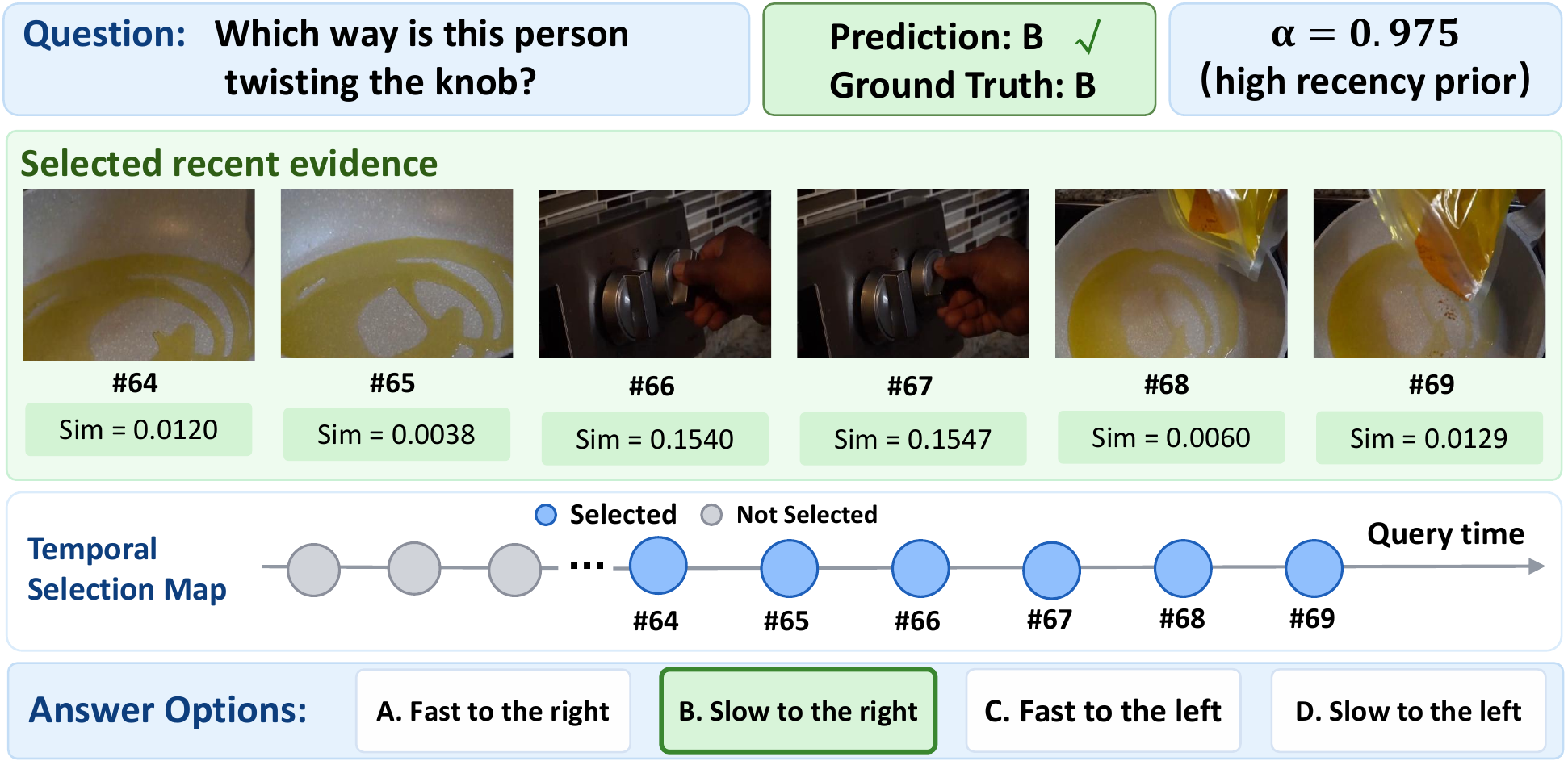}
        \caption{
            \textbf{Staying recent when local evidence is sufficient.}
        }
        \label{fig:caseB}
    \end{subfigure}

        \vspace{-1mm}

        \caption{
            \textbf{Qualitative examples of adaptive evidence allocation.}
            \textbf{(a)} When earlier observations are more relevant to the query,
            WRWS retrieves historical evidence while keeping recent context.
            \textbf{(b)} When the relevant action occurs near the query time,
            WRWS shifts the evidence budget toward recent observations.
            WRWS answers both queries correctly.
        }

        \vspace{-1mm}
        
    \label{fig:qualitative_cases}
\end{figure*}

\begin{figure}[t]
    \centering

    \begin{subfigure}[t]{0.49\linewidth}
        \centering
        \includegraphics[width=\linewidth]{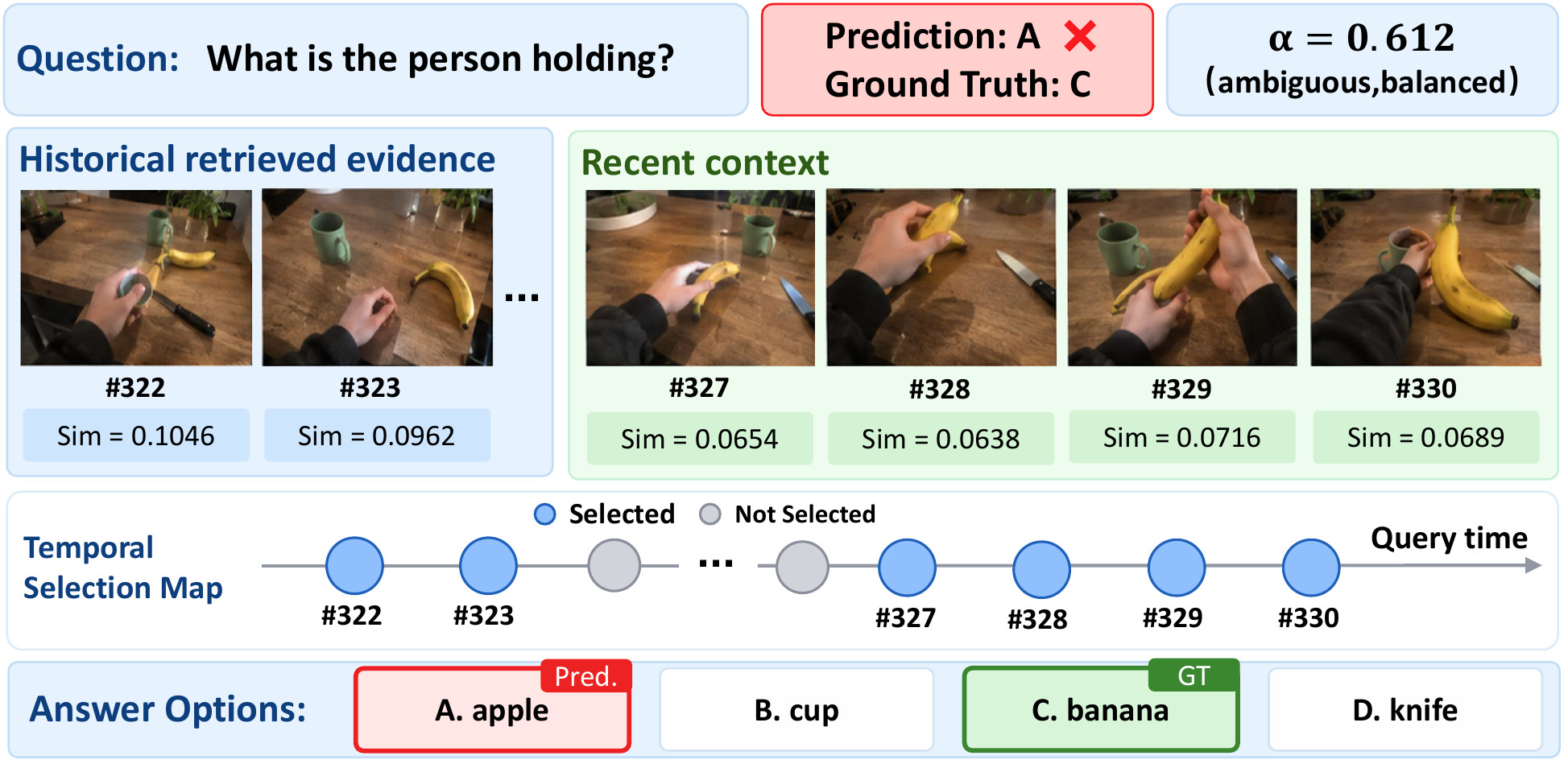}
        \caption{
            \textbf{Failure despite relevant selected evidence.}
        }
        \label{fig:caseC}
    \end{subfigure}
    \hfill
    \begin{subfigure}[t]{0.49\linewidth}
        \centering
        \includegraphics[width=\linewidth]{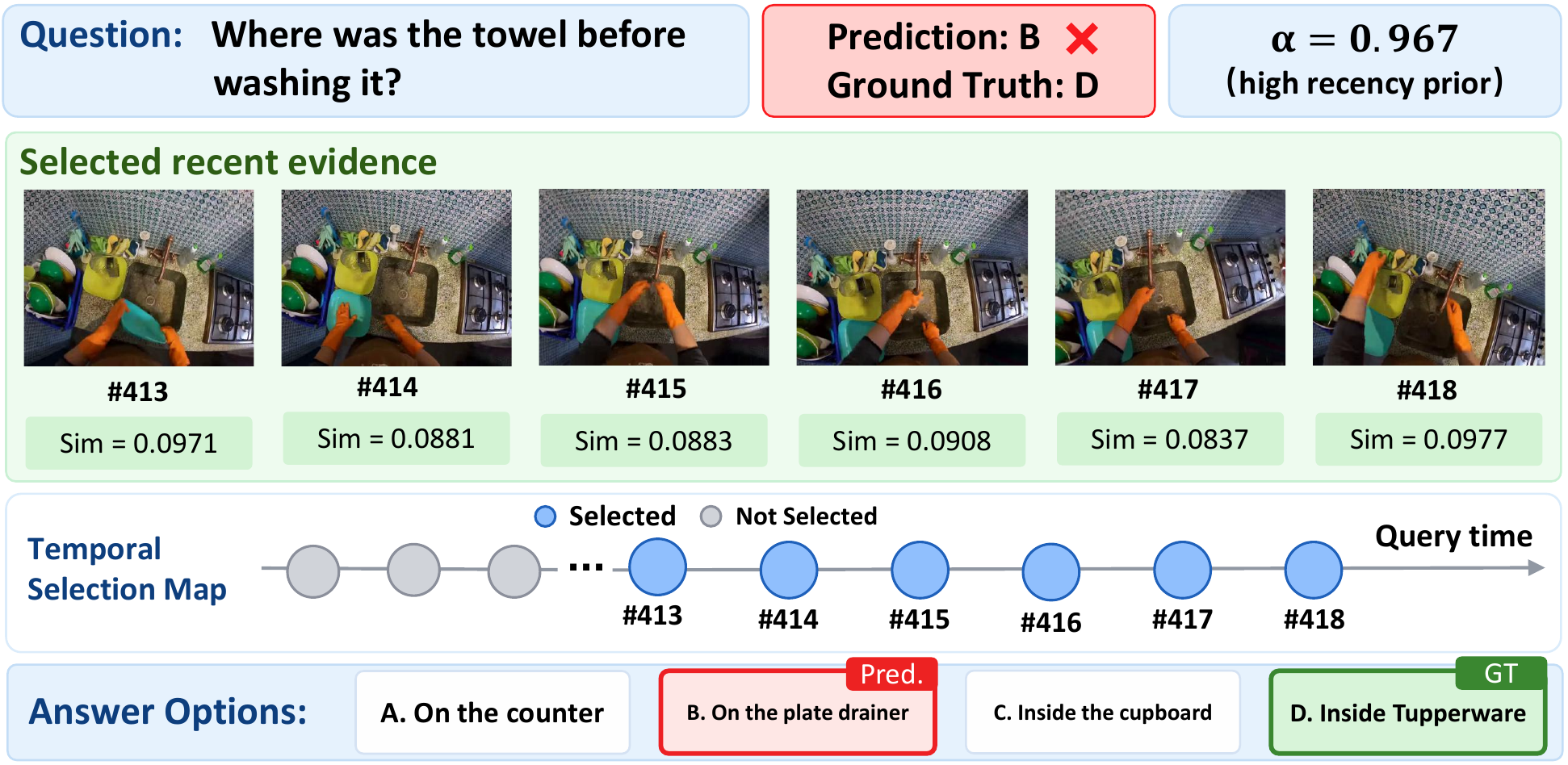}
        \caption{
            \textbf{Failure when earlier evidence is missed.}
        }
        \label{fig:caseD}
    \end{subfigure}
    
    \vspace{-1mm}

    \caption{
        \textbf{Qualitative failure cases.}
        \textbf{(a)} WRWS selects both earlier and recent observations in which the banana is visible, yet the target Video-LLM predicts ``apple'' instead of ``banana''.
        This suggests that evidence selection alone does not eliminate downstream visual recognition or reasoning errors.
        \textbf{(b)} When WRWS concentrates its budget on recent observations, the earlier evidence needed to recover the towel's previous location is not selected, leading to an incorrect answer.
    }
    \label{fig:failure_cases}
    \vspace{-5mm}
\end{figure}

We further examine representative examples to understand how WRWS allocates evidence across time and where its remaining errors arise.

Figs.~\ref{fig:caseA} and~\ref{fig:caseB} illustrate two successful allocation patterns.
In Fig.~\ref{fig:caseA}, several earlier observations receive stronger semantic similarity than the recent context and are retained together with recent frames.
This allows WRWS to recover historical evidence without discarding local context near the query time.
Fig.~\ref{fig:caseB} shows the opposite regime: the evidence needed to answer the query is already contained in the recent observations, and WRWS keeps the allocation concentrated near the query time.
Together, these examples illustrate the intended behavior of the adaptive semantic--recency trade-off rather than committing to a fixed temporal policy.

The failure cases reveal two different sources of error.
In Fig.~\ref{fig:caseC}, answer-relevant visual evidence is already present in the selected observations, yet the target Video-LLM still predicts an incorrect option.
The failure therefore cannot be explained solely by evidence selection, and instead suggests a limitation in downstream visual recognition or reasoning.
By contrast, Fig.~\ref{fig:caseD} exposes a selection failure.
The query requires information from an earlier stage of the video, while the selected set is dominated by recent observations around the current washing activity.
The relevant historical evidence is consequently absent from the target model's input.
These cases show that uncertainty-adaptive allocation improves temporal evidence selection but does not eliminate errors arising from either evidence allocation or the reasoning capability of the underlying Video-LLM.

\section{Limitations and Future Work}
\label{app:limitations}

WRWS currently retains lightweight retrieval representations for the observed history, so the retrieval-side cache grows with stream length. 
While these representations are substantially more compact than target-model visual states and can be offloaded from GPU memory, combining WRWS with cache compression or eviction could further improve scalability to very long streams. 
In addition, the current allocation policy estimates retrieval uncertainty from the similarity distribution of an external encoder. 
Future work may explore improved uncertainty calibration and richer retrieval signals to further refine adaptive evidence allocation.

\end{document}